\documentclass[runningheads]{llncs}
\usepackage[utf8]{inputenc}
\usepackage[T1]{fontenc}

\usepackage[american]{babel}

\usepackage[sort&compress,numbers]{natbib}
\makeatletter
\renewcommand\@biblabel[1]{#1.}
\makeatother
\usepackage{mathtools}
\usepackage{amssymb}
\usepackage{bm}
\usepackage{siunitx}
\usepackage{booktabs}
\usepackage{tabu}
\usepackage{csvsimple}
\usepackage{multicol}
\usepackage{multirow}
\usepackage{enumitem}
\usepackage{xstring}
\usepackage{tikz}
\usetikzlibrary{backgrounds,calc}
\usepackage{pgfplots}
\pgfplotsset{compat=1.5}
\pgfplotsset{
    cycle list={t_blue\\t_red\\t_green\\t_yellow\\},
}
\usepgfplotslibrary{fillbetween}
\usepackage{intcalc}
\usepackage{graphicx}
\usepackage[misc]{ifsym}

\usepackage{hyperref}
\hypersetup{
    plainpages=false,           
    pdfborder={0 0 0},          
    breaklinks=true,            
    bookmarksnumbered=true,     %
    bookmarksopen=true,         %
	hypertexnames=true,         %
}

\usepackage[capitalise,nameinlink]{cleveref}
\crefname{ax}{axiom}{axioms}
\usepackage{acro}
\acsetup{format/first-long=\slshape} 
\acsetup{barriers/use=true}
\acsetup{barriers/reset=true}
\acsetup{single}

\newcommand{\dac}[3]{\DeclareAcronym{#1}{short = #2, long = #3}}

\DeclareMathOperator*{\argmin}{arg\,min}

\newcommand{\bftab}{\fontseries{b}\selectfont}
\newcommand*{\condbold}[3][]{\ifthenelse{\equal{#2}{1}#1}{{\bftab #3}}{#3}}
\newcommand*{\rawres}[3]{\ifthenelse{\equal{#1}{}}{-}{\condbold{#3}{#1}}}
\pgfplotscreateplotcyclelist{accrejPlotColors}{%
t_green\\%
t_blue\\%
t_red\\%
}
\newcommand*{\addSE}[3]{
    \addplot[name path=#2upper, draw=none] table[x=p, y expr=\thisrow{#2Mean}+\thisrow{#2SE}, col sep=comma] {#1};
    \addplot[name path=#2lower, draw=none] table[x=p, y expr=\thisrow{#2Mean}-\thisrow{#2SE}, col sep=comma] {#1};
    \addplot[fill=#3, fill opacity=0.2] fill between[of=#2upper and #2lower];
}

\usepackage{colortbl}
\definecolor{t_gray}{HTML}{888888}
\definecolor{t_blue}{HTML}{355fb3}
\definecolor{t_red}{HTML}{b33535}
\definecolor{t_green}{HTML}{3bb335}
\definecolor{t_yellow}{HTML}{b39735}
\definecolor{t_darkgray}{HTML}{454545}
\definecolor{t_darkblue}{HTML}{1e3666}
\definecolor{t_darkgreen}{HTML}{22661e}
\definecolor{t_darkred}{HTML}{661e1e}
\definecolor{t_darkyellow}{HTML}{66571e}
\definecolor{t_lightblue}{HTML}{8ea7d7}
\definecolor{t_lightred}{HTML}{dc8989}
\definecolor{t_lightgreen}{HTML}{8ddc89}

\dac{ml}{ML}{machine learning}
\dac{nlp}{NLP}{natural language processing}
\dac{nn}{NN}{neural network}

\dac{mlp}{MLP}{multilayer perceptron}
\dac{rnn}{RNN}{Recurrent Neural Network}
\dac{svm}{SVM}{Support Vector Machine}
\dac{cnn}{CNN}{Convolutional Neural Network}
\dac{gnn}{GNN}{Graph Neural Network}
\dac{gcn}{GCN}{Graph Convolutional Network}
\dac{gat}{GAT}{Graph Attention Network}
\dac{gin}{GIN}{Graph Isomorphism Network}
\dac{ggpn}{GGPN}{Generalized Graph Posterior Network}
\dac{gpn}{GPN}{Graph Posterior Network}
\dac{cuqgnn}{CUQ-GNN}{Committe-based Uncertainty Quantification Graph Neural Network}
\dac{llop}{LLOP}{log-linear opinion pooling}
\dac{lop}{LOP}{linear opinion pooling}
\dac{lopgpn}{LOP-GPN}{linear opinion pooled graph posterior network}
\dac{au}{AU}{aleatoric uncertainty}
\dac{eu}{EU}{epistemic uncertainty}
\dac{tu}{TU}{total uncertainty}
\dac{uq}{UQ}{uncertainty quantification}
\dac{appnp}{APPNP}{approximate personalized propagation of neural predictions}
\dac{postnet}{PostNet}{Posterior Network}
\dac{ce}{CE}{cross-entropy}
\dac{uce}{UCE}{uncertain cross-entropy}
\dac{ppr}{PPR}{personalized Page-Rank}
\dac{arc}{ARC}{accuracy-rejection curve}
\dac{ood}{OOD}{out-of-distribution}
\dac{id}{ID}{in-distribution}
\dac{auroc}{AUC-ROC}{area under the receiver operating characteristic curve}
\dac{nf}{NF}{Normalizing Flow}
\dac{pdf}{PDF}{probability density function}
\dac{kl}{KL}{Kullback-Leibler}
\dac{elu}{ELU}{Exponential Linear Unit}
\dac{gkde}{GKDE}{Graph-based Kernel Dirichlet distribution Estimation}

\newcommand{\corr}{(\Letter)}

\begin{document}

\title{CUQ-GNN: Committee-based Graph Uncertainty Quantification using Posterior Networks}
\titlerunning{CUQ-GNN: Committee-based Graph UQ using PostNets}
\toctitle{CUQ-GNN: Committee-based Graph Uncertainty Quantification using Posterior Networks}

\author{Clemens Damke\inst{1} \corr
\and
Eyke Hüllermeier\inst{1,2}}
\tocauthor{Clemens~Damke,Eyke~Hüllermeier}

\authorrunning{C. Damke and E. Hüllermeier}

\institute{Institute of Informatics, LMU Munich, Germany \email{\{clemens.damke,eyke\}@ifi.lmu.de}
\and
Munich Center for Machine Learning (MCML), Germany}

\maketitle              

\begin{abstract}

In this work, we study the influence of domain-specific characteristics when defining a meaningful notion of predictive uncertainty on graph data.
Previously, the so-called \ac{gpn} model has been proposed to quantify uncertainty in node classification tasks.
Given a graph, it uses \acp{nf} to estimate class densities for each node independently and converts those densities into Dirichlet pseudo-counts, which are then dispersed through the graph using the \ac{ppr} algorithm.
The architecture of \acp{gpn} is motivated by a set of three axioms on the properties of its uncertainty estimates.
We show that those axioms are not always satisfied in practice and therefore propose the family of \acp{cuqgnn}, which combine standard \acp{gnn} with the \acs{nf}-based uncertainty estimation of \acp{postnet}.
This approach adapts more flexibly to domain-specific demands on the properties of uncertainty estimates.
We compare \ac{cuqgnn} against \ac{gpn} and other uncertainty quantification approaches on common node classification benchmarks and show that it is effective at producing useful uncertainty estimates.

\end{abstract}
\acbarrier%

\section{Introduction}

In machine learning systems, particularly those where safety is important, accurately quantifying prediction uncertainty is paramount.
One source of predictive uncertainty is the inherent stochasticity of the data-generating process, which is referred to as \acf{au} and cannot be reduced by sampling additional data.
For example, when tossing a fair coin, the outcome is uncertain, and this uncertainty is of purely aleatoric nature.
\Acf{eu}, on the other hand, arises from a lack of knowledge about the data-generating process.
It can be reduced by collecting more data and should vanish in the limit of infinite data~\citep{hullermeier2021}.
For example, the lack of knowledge about the bias of a coin is of epistemic nature, and it increases the (total) uncertainty about the outcome of a coin toss. This uncertainty, however, can be reduced by tossing the coin repeatedly and estimating the bias from the outcomes.

In the context of graph data, the structural information is an additional contributing factor to the uncertainty, making \ac{uq} particularly challenging.
In this paper, we will focus specifically on the problem of \ac{uq} for (semi-supervised) node classification.
Applications of this problem include, for example, the classification of documents in citation networks~\citep{sen2008,bojchevski2018}, or the classification of users or posts in social networks~\citep{shu2017}.

Recently, \citet{stadler2021} have proposed \acp{gpn} as an approach to \ac{uq} for node classification.
In this paper, we provide a novel perspective on \ac{gpn} motivated by the field of risk and decision analysis~\citep{clemen1999,clemen2007}.
More specifically, we show that \ac{gpn} can be interpreted as an opinion pooling model employing the so-called \ac{llop} scheme for structure-aware \ac{uq} (\cref{sec:uq}).
Motivated by this interpretation, we describe the limitations of \ac{gpn} and its axiomatic approach to \ac{uq} in general (\cref{seq:cuqgnn}).
To address those limitations, we introduce the family of \ac{cuqgnn} models, which is motivated by the notion of \emph{behavioral pooling}, combining standard \acp{gnn} with the \ac{postnet}.
Then, we compare our behavioral \ac{cuqgnn} model against the axiomatic \ac{gpn} model (\Cref{sec:eval}).
The effectiveness of \ac{cuqgnn} is demonstrated on multiple common node classification benchmarks.

\section{Uncertainty Quantification}\label{sec:um}

There are different formalizations of uncertainty in the literature on \ac{uq}.
Depending on the desired properties of the uncertainty measure, different notions may be more or less suitable.
We evaluate the adequacy of a measure of uncertainty through two lenses:
\begin{enumerate}
    \item Its adherence to a set of axioms~\citep{pal1993,bronevich2008,wimmer2023,sale2023a}.
    \item Its performance on a predictive task, such as outlier detection~\citep{charpentier2020}.
\end{enumerate}
Given our focus on \ac{uq} for the node classification setting, we start with a brief overview of uncertainty measures for classification tasks.

\subsection{Entropy-based Uncertainty Measures}\label{sec:um:entropy}

Predictions in $K$-class classification tasks are commonly represented in terms of probability distributions $\theta = (\theta_1, \ldots , \theta_K)\in \Delta_K$, where $\Delta_K$ represents the unit $(K-1)$-simplex, and $\theta_k$ denotes the probability of the $k$-th class. Thus, the true outcome (materialized class label) $Y$ remains uncertain and can be seen as a matter of chance (aleatoric uncertainty). In addition, the prediction $\theta$ itself is normally uncertain, too. This (epistemic) uncertainty of the learner can be represented through a second-order probability distribution $Q$ on $\Delta_K$.
Consequently, the true distribution over the $K$ classes is viewed as a random variable $\Theta \sim Q$.
Given a second-order distribution $Q$, its expectation is
\begin{equation}
    \bar{\theta} \coloneqq \mathbb{E}_{Q}[\Theta] = \int_{\Delta_K} \theta \, \mathrm{d}Q(\theta) \, . \label{eq:q-exp}
\end{equation}
The \acf{tu} regarding the outcome $Y$ is quantified by the Shannon entropy of $\bar{\theta}$:
\begin{equation}
    \mathrm{TU}(Q) \coloneqq H(\mathbb{E}_{Q}[\theta]) = -\sum_{k=1}^{K} \bar{\theta}_k \log \bar{\theta}_k . \label{eq:tu}
\end{equation}
Furthermore, a breakdown of this uncertainty into aleatoric and epistemic components can be achieved through a well-established result from information theory, which states that entropy is the sum of conditional entropy and mutual information~\citep{kendall2017,depeweg2018}.
This result suggests a quantification of \acf{au} as conditional entropy (of the outcome $Y$ given the first-order distribution $\Theta$):
\begin{equation}
    \mathrm{AU}(Q) \coloneqq \mathbb{E}_{Q} \left[ H(\Theta) \right] = -\smashoperator{\int_{\Delta_K}} \sum_{k=1}^{K} \theta_k \log \theta_k \, \mathrm{d}Q(\theta) \, . \label{eq:au}
\end{equation}
Moreover, the \acf{eu} can then be defined as the difference between \ac{tu} and \ac{au}:
\begin{align}
    \mathrm{EU}(Q) & \coloneqq \mathrm{TU}(Q) - \mathrm{AU}(Q)
    = I(Y; \Theta) = \mathbb{E}_{Q}[D_{\mathrm{KL}}(\Theta \| \bar{\theta})] \, ,  \label{eq:eu}
\end{align}
where $I(\cdot; \cdot)$ denotes mutual information and $D_{\mathrm{KL}}(\cdot \| \cdot)$ the \ac{kl} divergence.
Another entropy-based approach to quantify \ac{eu} is via the differential entropy of the second-order distribution $Q$~\citep{malinin2018,kotelevskii2023}:
\begin{equation}
    \mathrm{EU}_{\textrm{SO}}(Q) \coloneqq H(Q) = -\int_{\Delta_K} \log Q(\theta) \, \mathrm{d}Q(\theta). \label{eq:eu-so-entropy}
\end{equation}
We will refer to this notion of \ac{eu} as \emph{second-order epistemic uncertainty}.
Note that the differential entropy of $Q$ can be negative, with $-\infty$ representing a state of no uncertainty, i.e., a Dirac measure.
However, some axiomatic characterizations of uncertainty assume that a state of no uncertainty is represented by an uncertainty of zero~\citep{wimmer2023}; $\mathrm{EU}_{\textrm{SO}}$ is therefore not without controvery.
Apart from the entropy-based measures we just described, uncertainty is also often quantified in terms of other concentration measures, such as variance~\citep{sale2023a,duan2024}, confidence or Dirichlet pseudo-counts.
We will now briefly review the latter two notions of uncertainty.

\subsection{Least-confidence and Count-based Uncertainty Measures}\label{sec:um:confidence}

Given a second-order distribution $Q$, an alternative notion of uncertainty is provided by the so-called \emph{least-confidence} of the expected distribution $\bar{\theta}$, defined as
$\mathrm{LConf}(Q) \coloneqq 1 - \max_k \bar{\theta}_k$.
Note the similarity of this measure to the \ac{tu} measure in \cref{eq:tu}; $\mathrm{LConf}(Q)$ can therefore be seen as a measure of \emph{total} uncertainty, too.
However, in the literature this measure is also used as a proxy for \emph{aleatoric} uncertainty~\citep{charpentier2020}.

Finally, if $Q$ is described by a Dirichlet distribution $\mathrm{Dir}(\bm{\alpha})$, where $\bm{\alpha} = (\alpha_1, \dots, \alpha_K)$ is a vector of pseudo-counts,
the sum $\alpha_0 = \sum_{k=1}^K \alpha_k$
describes how concentrated $Q$ is around the expected distribution $\bar{\theta}$.
Thus, the \ac{eu} encoded by a Dirichlet distribution $Q$ can be quantified by
$\mathrm{EU}_{\textrm{PC}}(Q) \coloneqq -\alpha_0$, which we will call \emph{pseudo-count-based epistemic uncertainty}~\citep{charpentier2020,huseljic2021,kopetzki2021}.

\section{Posterior and Graph Posterior Networks}\label{sec:uq}

As just described, uncertainty can be formalized in various ways, and the selection of an uncertainty measure hinges upon the specific criteria it should satisfy.
In graph-related contexts, an additional element contributing to uncertainty, namely structural information, requires formalization.
\Citet{stadler2021} propose an axiomatic approach to account for structure-induced uncertainty, called \acf{gpn}.
As mentioned in the introduction, \acp{gpn} are essentially a combination of \acp{postnet}~\citep{charpentier2020} and the \ac{appnp} node classification model~\citep{gasteiger2018}.
We begin with a review of the \acp{postnet} and \acp{gpn} and the notions of uncertainty they provide.

\subsection{Posterior Networks}\label{sec:uq:postnet}

A \ac{postnet} is an \emph{evidential deep learning} classification model~\citep{sensoy2018}, quantifying predictive uncertainty via a second-order distribution $Q$, learned through a second-order loss function $L_2$.
A standard (first-order) loss function $L_1 : \Delta_K \times \mathcal{Y} \to \mathbb{R}$ takes a predicted first-order distribution $\hat{\theta} \in \Delta_K$ and an observed ground-truth label $y \in \mathcal{Y}$ as input (where $\mathcal{Y}$ denotes the set of classes); the \acf{ce} loss is a common example of such a first-order loss function.
Similarly, a second-order loss $L_2$ takes a second-order distribution $Q$, i.e., a distribution over $\Delta_K$, as input, to which it again assigns a loss in light of an observed label $y \in \mathcal{Y}$.
\Ac{postnet} uses the so-called \acf{uce} loss~\citep{bilos2019}, which is defined as
\begin{align}
    L_2(Q, y) & \coloneqq \mathbb{E}_{Q} \left[ \mathrm{CE}(\Theta, y) \right]
    = -\int_{\Delta_K} \log P(y\, |\, \theta) \, \mathrm{d}Q(\theta) \, . \label{eq:uce}
\end{align}
However, directly minimizing a second-order loss, like the \ac{uce} loss, presents challenges, as the minimum is attained when $Q$ is a Dirac measure concentrating all probability mass on $\theta^* = {\argmin}_{\theta \in \Delta_K} \mathrm{CE}(\theta, y)$~\citep{bengs2022}.
Consequently, for the notions of \ac{eu} we discussed in \cref{sec:um}, the optimal $Q^*$ will have no \ac{eu}, i.e., $\mathrm{EU} = 0$~(\cref{eq:eu}) and $\mathrm{EU}_{\textrm{PC}} = -\infty$~(\cref{sec:um:confidence}).
To mitigate this issue, a regularization term, typically the differential entropy of $Q$, is added to the second-order loss function, incentivizing a $Q$ that is less concentrated.
Whether one can obtain a faithful representation of \acl{eu} has been generally questioned by \citet{bengs2023}.
One should therefore be cautious when interpreting the \ac{eu} estimates of evidential deep learning models, such as \ac{postnet}.
We will not attempt to interpret uncertainty estimates in a quantitative manner but rather focus on the question of whether they are qualitatively meaningful, e.g., by considering whether anomalous or noisy instances can be identified via their uncertainty.

\Ac{postnet} models the second-order distribution $Q$ as a Dirichlet distribution $\mathrm{Dir}(\mathbf{\alpha})$, where $\mathbf{\alpha} = (\alpha_1, \dots, \alpha_K)$ is a vector of pseudo-counts.
The predicted pseudo-counts $\alpha_k$ for a given instance $\mathbf{x}^{(i)} \in \mathcal{X}$ are defined as
\begin{equation}
    \alpha_k = 1 + \mu \cdot P \left(\mathbf{z}^{(i)}\,|\, y^{(i)} = k \right) \cdot P \left(y^{(i)} = k \right) ,
\end{equation}
where $\mathbf{z}^{(i)} = f(\mathbf{x}^{(i)}) \in \mathbb{R}^H$ is a latent neural network embedding of $\mathbf{x}^{(i)}$ and $\mu \in \mathbb{R}$ a so-called \emph{certainty budget}, determining the highest attainable pseudo-count for a given instance.
The class-conditional probability $P(\mathbf{z}^{(i)}\,|\, k)$ by a normalizing flow model for the class $k$ estimates the density of the instance.
Overall, the \ac{postnet} model therefore consists of a neural network encoder model $f$ and $K$ normalizing flow models, one for each class.

\subsection{Graph Posterior Networks}\label{sec:uq:gpn}

\Acfp{gpn} extend the \ac{postnet} model to the node classification problem in graphs.
Let $G \coloneqq (\mathcal{V}, \mathcal{E})$ denote a graph, where $\mathcal{V}$ is a set of $N \coloneqq |\mathcal{V}|$ nodes and $\mathcal{E} \subseteq \mathcal{V}^2$ the set of edges.
The adjacency matrix of $G$ is denoted by $\mathbf{A} = (A_{i,j}) \in {\{0,1\}}^{N \times N}$, where $A_{i,j} = 1$ iff $(v_i, v_j) \in \mathcal{E}$.
For simplicity, we also assume that $G$ is undirected, i.e., that $\mathbf{A}$ is symmetric.
For each node $v^{(i)} \in \mathcal{V}$ we have a feature vector $\mathbf{x}^{(i)} \in \mathbb{R}^D$ and a label $y^{(i)} \in \mathcal{Y}$.
The goal of the node classification task is to predict the label of each node in $\mathcal{V}$, given the graph structure and the node features.

\Acp{gpn} classify the nodes of a given graph by first making a prediction for each node $v^{(i)}$ solely based on its features $\mathbf{x}^{(i)}$ using a standard \ac{postnet} model, i.e., without considering the graph structure.
The predicted feature-based pseudo-count vectors $\mathbf{\alpha}^{\mathrm{ft},(i)}$ for each vertex $v^{(i)}$ are then dispersed through the graph via a \ac{ppr} matrix $\mathbf{\Pi}^{\mathrm{PPR}} \in \mathbb{R}^{N \times N}$ as follows:
\begin{align}
    \mathbf{\alpha}^{\mathrm{agg},(i)} &\coloneqq \sum_{v^{(j)} \in \mathcal{V}} \mathbf{\Pi}^{\mathrm{PPR}}_{i,j} \mathbf{\alpha}^{\mathrm{ft},(j)} \label{eq:gpn-alpha} \\
    \text{where } \mathbf{\Pi}^{\mathrm{PPR}} &\coloneqq {\left(\varepsilon \mathbf{I} + (1 - \varepsilon) \mathbf{\hat{A}}\right)}^L \label{eq:gpn-ppr}
\end{align}
Here, $\mathbf{I}$ is the identity matrix, $\varepsilon \in (0,1]$ the so-called \emph{teleport probability}, and $\mathbf{\hat{A}} \coloneqq \mathbf{A} \mathbf{D}^{-1}$ the normalized (random-walk) adjacency matrix, with $\mathbf{D} \coloneqq \mathrm{diag}(\mathbf{A} \mathbf{1})$ being the degree matrix of $G$.
For large $L$, $\mathbf{\Pi}^{\mathrm{PPR}}$ approximates the personalized page-rank matrix of the graph via power iteration.
\Citet{gasteiger2018} proposed this page-rank inspired information dispersion scheme for the node classification task, which they refer to as \ac{appnp}.
The main difference between \ac{appnp} and \ac{gpn} is that \ac{appnp} disperses (first-order) class probability vectors $\theta^{\mathrm{ft},(i)}$ for each node $v^{(i)}$, whereas \ac{gpn} disperses pseudo-count vectors $\mathbf{\alpha}^{\mathrm{ft},(i)}$.

To justify this pseudo-count dispersion scheme, \citet{stadler2021} propose the following three axioms on how the structural information in a graph should influence the uncertainty of a model's predictions:
\begin{enumerate}[label=\textbf{A\arabic*}]
    \item A node's prediction should only depend on its own features in the absence of network effects.
        A node with features more different from training features should have a higher uncertainty. \label[ax]{ax:gpn-ft}
    \item All else being equal, if a node $v^{(i)}$ has a lower epistemic uncertainty than its neighbors in the absence of network effects, the neighbors' predictions should become less epistemically uncertain in the presence of network effects. \label[ax]{ax:gpn-eu}
    \item All else being equal, if a node $v^{(i)}$ has a higher aleatoric uncertainty than its neighbors in the absence of network effects, the neighbors' predictions should become more aleatorically uncertain in the presence of network effects.
        Further, the aleatoric uncertainty of a node in the presence of network effects should be higher if the predictions of its neighbors in the absence of network effects are more conflicting. \label[ax]{ax:gpn-au}
\end{enumerate}
\Citet{stadler2021} show the validity of those axioms if \ac{au} is defined as the least-confidence $\mathrm{LConf}$ and \ac{eu} as the negative sum of the pseudo-counts $\mathrm{EU}_{\mathrm{PC}}$ (\cref{sec:um:confidence}).
Using those definitions, the validity of the axioms follows from the fact that $\mathbf{\alpha}^{\mathrm{agg},(i)}$ is effectively a weighted average of the pseudo-counts of the (indirect) neighbors of $v^{(i)}$, with high weights for close neighbors and low weights for more distant ones.
The \ac{gpn} axioms are motivated by two assumptions, namely, \emph{network homophily} and the \emph{irreducibility of conflicts}.

First, \emph{network homophily} refers to the assumption that an edge implies similarity of the connected nodes; more specifically, in the context of \acp{gpn}, connected nodes should have similar second-order distributions, and thereby similar predictive uncertainties.
This is a common assumption shared by many \ac{gnn} architectures and based on the idea of repeatedly summing or averaging the features of each node's neighbors~\citep{kipf2017,xu2018}.
As already remarked by \citet{stadler2021}, non-homophilic graphs are not properly dealt with by \acp{gpn}, nor by other \ac{gnn} architectures in general~\citep{zhu2020}.
Nonetheless, since edges are typically used to represent some form of similarity, the homophily assumption is often reasonable.

Second, \emph{irreducibility of conflicts} refers to the assumption that conflicting predictions cannot be resolved by aggregating the predictions of the conflicting nodes.
\Cref{fig:gpn-aggregation} illustrates the implications of this assumption for binary node classification; there, without network effects, node \textsf{A} is very confident that its probability of belonging to the positive class is high, whereas node \textsf{B} is very confident that its probability of belonging to that class is low.
Thus, both nodes make conflicting predictions while both having a low \ac{au} and a low \ac{eu}.
Due to the homophily assumption, a consensus has to be found between the two conflicting predictions.
As described in \cref{ax:gpn-au}, a \ac{gpn} will do this by increasing the \ac{au} of the aggregated prediction while keeping the \ac{eu} low.
\Citet{stadler2021} argue that this is reasonable because such a conflict is inherently irreducible and should therefore be reflected in the \acl{au} of the aggregated prediction.

\subsection{GPN as an Opinion Pooling Model}\label{sec:uq:gpn-voting}

\begin{figure}[t]
    \centering
    \includegraphics[width=0.7\linewidth]{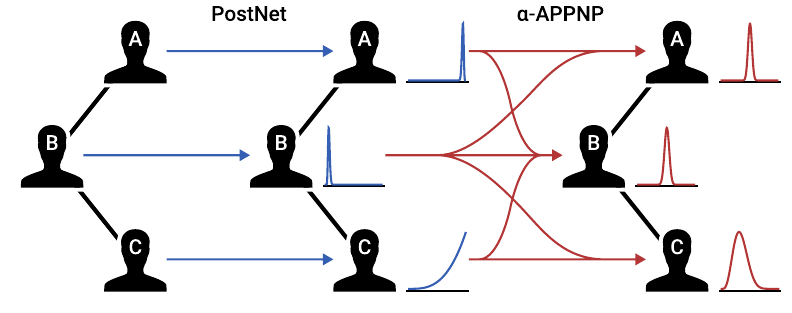}
    \caption{Illustration of how \acs*{gpn} aggregates the predictions (opinions) of different nodes (agents).}\label{fig:gpn-aggregation}
\end{figure}
In addition to the axiomatic motivation for the averaging of pseudo-counts, \acp{gpn} can also be interpreted from the perspective of \emph{opinion pooling}.
The question of how to pool, or aggregate, the opinions of multiple agents is studied in the fields of social choice theory~\citep{arrow1951} and decision and risk analysis~\citep{clemen1999,clemen2007}.
In the context of node classification, the agents are the nodes of the graph, and their opinions are the predicted second-order Dirichlet distributions.
\Cref{fig:gpn-aggregation} illustrates how \ac{gpn} aggregates the distributions at different nodes.
Let $Q^{\mathrm{ft},(i)} = \mathrm{Dir}(\alpha^{\mathrm{agg},(i)})$ be the distribution at node $v^{(i)}$ before dispersion and $Q^{\mathrm{agg},(i)} = \mathrm{Dir}(\alpha^{\mathrm{agg},(i)})$ the distribution after dispersion.
We can then express the aggregated distribution $Q^{\mathrm{agg},(i)}$ as follows:
\begin{align}
    Q^{\mathrm{agg},(i)}(\theta) &= \frac{1}{B(\alpha^{\mathrm{agg},(i)})} \prod_{k=1}^K \theta_k^{\alpha^{\mathrm{agg},(i)}_k - 1}
    = \frac{1}{B(\alpha^{\mathrm{agg},(i)})} \prod_{k=1}^K \theta_k^{\left(\sum_{j = 1}^N \mathbf{\Pi}^{\mathrm{PPR}}_{i,j} \mathbf{\alpha}^{\mathrm{ft},(j)}_k\right) - 1} \nonumber \\
    &= \underbrace{\frac{\prod_{j=1}^N {\left(B(\alpha^{\mathrm{ft},(j)})\right)}^{\mathbf{\Pi}^{\mathrm{PPR}}_{i,j}}}{B(\alpha^{\mathrm{agg},(i)})}}_{Z^{-1}} \prod_{j=1}^N {\left(\frac{1}{B(\alpha^{\mathrm{ft},(j)})} \prod_{k=1}^K \theta_k^{\mathbf{\alpha}^{\mathrm{ft},(j)}_k - 1}\right)}^{\mathbf{\Pi}^{\mathrm{PPR}}_{i,j}} \nonumber \\
    &= \frac{1}{Z} \prod_{j=1}^N {\left(Q^{\mathrm{ft},(i)}(\theta)\right)}^{\mathbf{\Pi}^{\mathrm{PPR}}_{i,j}}
    \ \text{with}\ Z = \int_{\Delta_K} \prod_{j=1}^K {\left(Q^{\mathrm{ft},(i)}(\theta)\right)}^{\mathbf{\Pi}^{\mathrm{PPR}}_{i,j}} \mathrm{d}\theta \label{eq:gpn-agg-density}
\end{align}
Here, $B(\cdot)$ denotes the (multivariate) Beta function and $Z$ a normalization constant.
This formulation of the aggregated distribution $Q^{\mathrm{agg},(i)}$ in terms of the distributions $Q^{\mathrm{ft},(j)}$ of the neighbors of $v^{(i)}$ is called \acf{llop}~\citep{genest1984,genest1986,clemen2007,koliander2022}.
\Ac{llop} is a natural opinion pooling scheme which is typically motivated by the fact that it satisfies so-called \emph{external Bayesianity}~\citep{genest1984}, i.e., applying a Bayesian update to the pooled opinion is equivalent to applying that update to all opinions before pooling.
In the context of Dirichlet opinions, external Bayesianity simply refers to the fact that adding a pseudo-count vector $\gamma \in \mathbb{R}_{\geq0}^K$ to the aggregated pseudo-count vector $\alpha^{\mathrm{agg},(i)}$ results in the same aggregated distribution as adding $\gamma$ to each of the pseudo-count vectors $\alpha^{\mathrm{ft},(j)}$ of the neighbors of $v^{(i)}$ and then aggregating.
Additionally, \citet{abbas2009} has shown that $Q^{\mathrm{agg},(i)}$ minimizes the expected \ac{kl} divergence $\sum_{j=1}^N \mathbf{\Pi}^{\mathrm{PPR}}_{i,j} D_{\mathrm{KL}}(Q^{\mathrm{agg},(i)} \| Q^{\mathrm{ft},(j)})$.

To summarize, \ac{gpn} can be understood as an opinion pooling model based on the \ac{llop} scheme.
Since there is a large variety of pooling schemes~\citep{koliander2022}, this raises the question of whether \ac{llop} is the most appropriate opinion pooling scheme for the node classification task.

\section{Committee-based Graph Uncertainty Quantification}\label{seq:cuqgnn}

In this section, we propose a new family of models, called \acf{cuqgnn}, which combines standard \acp{gnn} with the \ac{postnet} model.
We first discuss the validity of \ac{llop} scheme employed by \acp{gpn} and identify limitations of this approach.
Next, we show how those limitations can be addressed by so-called \emph{behavioral pooling} schemes.
Last, \ac{cuqgnn} is introduced as a concrete instantiation of this idea.

\subsection{The Irreducibility of Conflicts Assumption}\label{sec:cuqgnn:conflicts}

As explained in \cref{sec:uq:gpn}, \acp{gpn} are based on the assumption that conflicts between the predictions of neighboring nodes are irreducible.
In a recent work, we argued \citep{damke2024} that the validity of this \emph{irreducibility} assumption depends on the nature of the data-generating process which produces the graph whose nodes are to be classified.
We will now describe this argument since it serves as motivation the \ac{cuqgnn} model introduced in the next section.
As mentioned in the introduction, irreducibility in the context of \ac{uq} refers to uncertainty that cannot be reduced by additional information, which, in a machine learning context, essentially means sampling additional data~\citep{hullermeier2021}.
In the context of node classification, the data points are nodes; given a sample graph $G_N = (\mathcal{V}_N, \mathcal{E}_N)$ with $N$ vertices, increasing the sample size corresponds to sampling a graph $G_M = (\mathcal{V}_M, \mathcal{E}_M)$ with $M > N$ nodes from an assumed underlying data-generating distribution $P_\mathcal{G}$ over all graphs $\mathcal{G}$, such that $G_N$ is a subgraph of $G_M$.
The question of whether a conflict between a node $v^{(i)}$ and its neighbor $v^{(j)}$ is irreducible then becomes the question of whether the conflict persists in the limit of $M \to \infty$.
Let $\mathcal{N}_M(v^{(i)})$ be the set of neighbors of $v^{(i)}$ in $G_M$.
Assuming homophily, each node $v^{(\ell)}$ that is added to $\mathcal{N}_M(v^{(i)})$ should be \emph{similar} to $v^{(i)}$ with high probability.
Depending on the data-generating distribution $P_\mathcal{G}$, there are two possible scenarios:
\begin{enumerate}[label=\textbf{\arabic*.}]
    \item \textbf{Bounded degree sampling:}
        The neighborhood of $v^{(i)}$ does not grow with the sample size, i.e., $\mathbb{E}[|\mathcal{N}_M(v^{(i)})|] \in \mathcal{O}(1)$ as $M \to \infty$.
        In this situation, the conflict between $v^{(i)}$ and $v^{(j)}$ is indeed irreducible, as no additional data can be sampled to resolve the conflict.
        Thus, \cref{ax:gpn-au} of \ac{gpn} is reasonable, the irreducible uncertainty, i.e., \ac{au}, should increase with conflicting predictions.
    \item \textbf{Unbounded degree sampling:}
        The neighborhood of $v^{(i)}$ grows with the sample size, i.e., $\mathbb{E}[|\mathcal{N}_M(v^{(i)})|] \to \infty$ as $M \to \infty$.
        In this situation, the conflict is reducible, as it will eventually be resolved by the addition of more similar nodes to the neighborhood of $v^{(i)}$, which will outweigh the conflicting node $v^{(j)}$.
       Thus, the conflict resolution approach of \ac{gpn} is not reasonable; not \ac{au}, but rather the reducible uncertainty, \ac{eu}, should increase.
\end{enumerate}
We argue that unbounded degree sampling is more common in practical node classification tasks.
The Barabási-Albert model~\citep{barabasi1999} is a scale-free model which describes the growth behavior of many real-world graphs, such as the World Wide Web, social networks, or citation networks~\citep{albert2002,redner1998,wang2008}.
In this model, the expected degree of the $N$-th sampled node $v^{(i)}$ after $M-N$ additional nodes have been sampled is equal to $|\mathcal{N}_N(v^{(i)})| \cdot \sqrt{\frac{M}{N}}$.
Thus, for $M \to \infty$, the expected neighborhood size of a node goes to infinity.
\begin{figure}[t]
    \centering
    \includegraphics[width=0.7\linewidth]{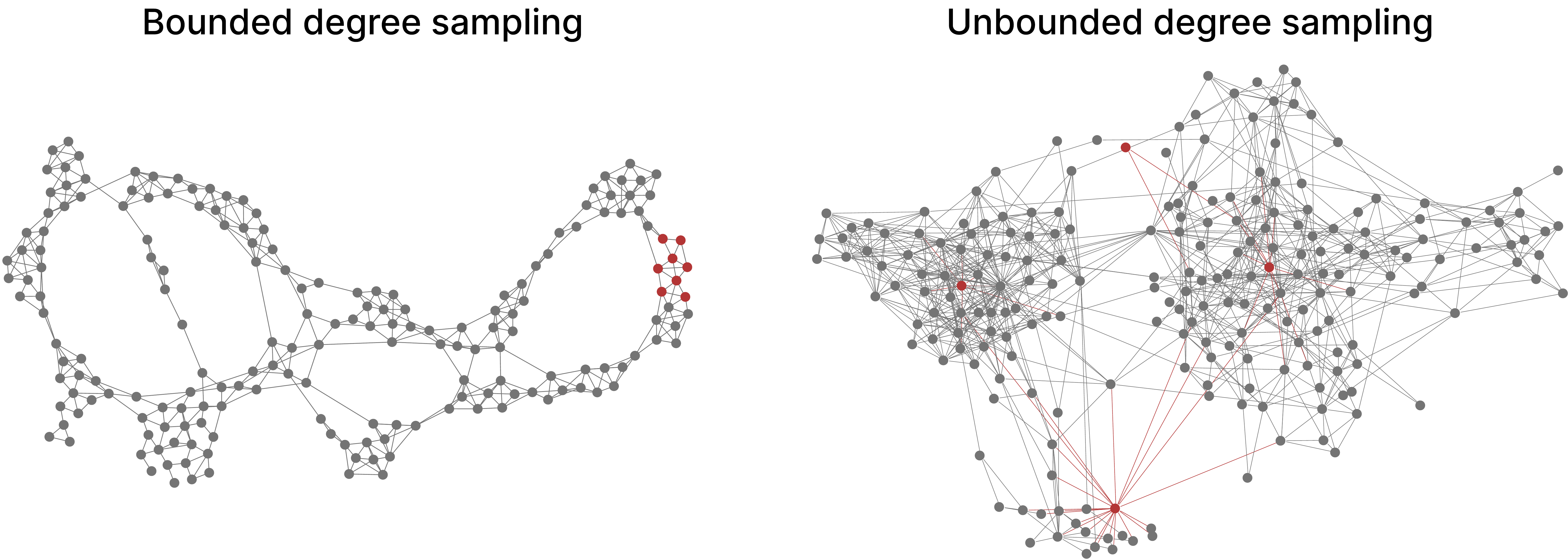}
    \caption{Examples for graphs obtained via bounded and unbounded degree sampling.}\label{fig:degree-sampling}
\end{figure}
Examples of domains in which the neighborhood sizes do not grow with the size of a graph are molecular graphs or lattice graphs, such as 3D models or images, which can be interpreted as grids of pixels (see~\cref{fig:degree-sampling}).
In those domains, node-level classification tasks are however less common, as one is typically interested in the classification of entire graphs, e.g., whether a given molecule is toxic or not.

To conclude, we argue that the axiomatic motivation for \acp{gpn} is oftentimes inappropriate.
Therefore, we propose a different approach to \ac{uq} for node classification which does not assume the irreducibility of conflicts from \cref{ax:gpn-au}.

\subsection{Resolving Conflicts via Behavioral Pooling}\label{sec:cuqgnn:behavioral}

The \ac{llop} scheme employed by \acp{gpn} is a so-called \emph{axiomatic pooling} approach~\citep{clemen2007}.
As described in the previous section, this axiomatic pooling scheme may not be appropriate if the domain does not satisfy the bounded degree sampling assumption.
For domains in which the unbounded degree sampling assumption holds, the so-called \acs{lopgpn} model~\citep{damke2024} has been shown to be a more appropriate choice.
As the name suggests, \ac{lopgpn} is a variant of \ac{gpn} that uses the axiomatic \ac{lop} scheme instead of \ac{llop}.
\Ac{lop} combines distributions by taking a weighted average of their densities, i.e., $Q^{\mathrm{agg},(i)} = \sum_{v^{(j)} \in \mathcal{V}} \mathbf{\Pi}^{\mathrm{PPR}}_{i,j} Q^{\mathrm{ft},(j)}$ (cf.~\cref{eq:gpn-agg-density}).
However, due to the inherent variability of real-world node classification tasks, it is difficult to determine a priori whether a given domain satisfies the bounded or unbounded degree sampling assumptions.
In practice, a domain might even exhibit a mixture of both sampling behaviors in different regions of a given graph.
Whether and when a local conflict in a graph is irreducible is a complex question that depends on the domain and its inherent data-generating distribution.
While the axiomatic motivations for \ac{gpn} and \ac{lopgpn} are appealing in theory, it is difficult to determine whether they are appropriate for a given problem.
Therefore, we argue that the appropriateness of a given \ac{uq} approach can only be meaningfully assessed by empirically comparing the quality and usefulness of the uncertainty estimates the produce.

This uncertainty about the appropriateness of different pooling schemes, motivates the idea of learning a pooling scheme from the data.
By giving up the theoretical guarantees about the properties of uncertainty estimates that axiomatic pooling schemes provide, a so-called \emph{behavioral pooling} approach can more flexibly adapt to the specific characteristics of a given task~\cite{clemen1999}.
While axiomatic pooling schemes combine the independent opinions of agents using a fixed aggregation rule, behavioral pooling schemes permit interactions between the agents.
All agents (nodes) that interact with a given agent $v^{(i)}$ can be interpreted as a group, or committee, $C^{(i)} \subseteq \mathcal{V}$ of agents which directly produces a joint opinion about $v^{(i)}$.
\begin{figure}[t]
    \centering
    \includegraphics[width=0.7\linewidth]{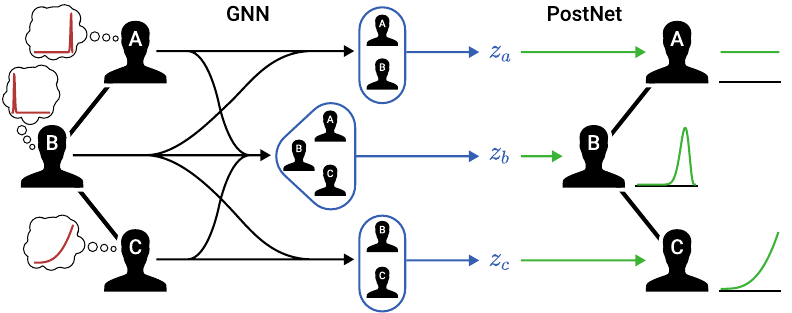}
    \caption{Illustration of how \acs*{cuqgnn} can flexibly resolve conflicts.}\label{fig:cuq-gnn}
\end{figure}
\Cref{fig:cuq-gnn} illustrates this idea.
In \cref{fig:gpn-aggregation}, the conflict between node \textsf{A} and \textsf{B} increases the pooled \ac{au}; by allowing \textsf{A} and \textsf{B} to exchange information, such a conflict can be resolved more flexibly, e.g., by predicting a uniform second-order distribution and thereby increasing \ac{eu}.

A behavioral pooling scheme is defined by two aspects:
First, the way the committees $C^{(i)}$ are constructed and, second, the way the agents within a committee interact.
A graph convolution, as used in \acp{gnn}, can be interpreted as such a behavioral pooling scheme, where $C^{(i)}$ is defined as a (indirect or higher-order) neighborhood of $v^{(i)}$ and the interaction between the agents is defined by the learned aggregation mechanism of the convolution operator.
There is a large variety of graph convolution operators in the literature, each defining a different way of constructing the committees and agent interactions.

Most common graph convolutions are based on the idea of neighborhood message-passing; a stack of $L$ graph convolution layers can be interpreted as a behavioral pooling scheme, where the committees $C^{(i)}$ are defined as the $L$-hop neighborhoods of $v^{(i)}$.
Examples of such neighborhood message-passing convolutions are \ac{appnp}~\citep{gasteiger2018}, \ac{gcn}~\citep{kipf2017} and \ac{gat}~\citep{velickovic2018}.
Alternatively, there are also so-called higher-order convolutions~\citep{maron2019,damke2020}, which do not operate directly on node features but on more complex substructures of a graph.
While the homophily assumption of \ac{gpn} is also (implicitly) made in many graph convolution operators, so-called heterogeneous graph convolutions~\citep{zhang2019a,wang2019} do not make this assumption.
Overall, different types of graph convolutions have been successfully employed on a wide range of tasks on graph data.

We propose a simple extension of graph convolutions to the \ac{uq} setting by combining them with the \ac{postnet} model~\citep{charpentier2020}.
We call this combination \emph{\ac{cuqgnn}}:
\begin{align}
    \textrm{CUQ-GNN}(\mathbf{X}, \mathbf{A}) = \textrm{PostNet}\left(\textrm{GNN}\left(h_{\mathit{enc}}(\mathbf{X}), \mathbf{A}\right) \cdot \mathbf{W}_{\mathit{lat}}\right) \label{eq:cuqgnn}
\end{align}
Here, $\mathbf{X} \in \mathbb{R}^{N \times d_{\mathit{in}}}$ is the input node feature matrix, $\mathbf{A} \in \mathbb{R}^{N \times N}$ the adjacency matrix, $h_{\mathit{enc}}: \mathbb{R}^{N \times d_{\mathit{in}}} \to \mathbb{R}^{N \times d_{\mathit{hid}}}$ a \ac{mlp}, $\textrm{GNN}: \mathbb{R}^{N \times d_{\mathit{hid}}} \times \mathbb{R}^{N \times N} \to \mathbb{R}^{N \times d_{\mathit{hid}}}$ a stack of graph convolution operators, $\mathbf{W}_{\mathit{lat}} \in \mathbb{R}^{d_{\mathit{hid}} \times d_{\mathit{lat}}}$ a latent embedding matrix and $\textrm{PostNet}: \mathbb{R}^{N \times d_{\mathit{lat}}} \to \mathbb{R}_{\geq 1}^{N \times K}$ a \ac{postnet} model.

By choosing an appropriate graph convolution operator, \ac{cuqgnn} can be applied to a wide range of node classification tasks.
Unlike \ac{gpn}, \ac{cuqgnn} neither assumes homophily nor the irreducibility of conflicts.
This flexibility comes at the cost of having no provable guarantees about the uncertainty estimates.
Nevertheless, due to the questionable real-world applicability of the axioms of \ac{gpn}, we argue that giving up (potentially) invalid axioms for a more flexible propagation of uncertainties through the graph is a justifiable trade-off.

\section{Evaluation}\label{sec:eval}

We assess the quality of the uncertainty estimates of \ac{cuqgnn} in two ways.
First, we compare the quality of uncertainty estimates of different \ac{cuqgnn} variants and \ac{gpn} using \acp{arc}.
Second, we compare the effectiveness of uncertainty estimates in detecting anomalous instances in two \ac{ood} settings.
All experiments are conducted on six common node classification benchmarks.%
\footnote{Implementation available at \url{https://github.com/Cortys/gpn-extensions}}

\subsection{Experimental Setup}\label{sec:eval:setup}
\paragraph{Datasets}
We use the following node classification benchmarks:
Three citation network datasets, namely, \textbf{CoraML}, \textbf{CiteSeer} and \textbf{PubMed}~\citep{mccallum2000,giles1998,getoor2005,sen2008,namata2012},
two co-purchase datasets, namely \textbf{Amazon Photos} and \textbf{Amazon Computers}~\citep{mcauley2015}
and the large-scale \textbf{OGBN Arxiv} dataset with about $170 \mathrm{k}$ nodes and over $2.3$ million edges~\citep{hu2020}.
Since OGBN Arxiv is presplit into train, validation and test sets, we use the provided splits.
The results for the other datasets are obtained by averaging over 10 random class-stratified splits of the node set with train/val/test sizes of $5\%$/$15\%$/$80\%$.

\paragraph{Models}
We evaluate three variants of \ac{cuqgnn}: \textbf{CUQ-PPR}, \textbf{CUQ-GCN} and \textbf{CUQ-GAT}.
CUQ-PPR uses \ac{appnp}~\citep{gasteiger2018}, CUQ-GCN uses \ac{gcn}~\citep{kipf2017} and CUQ-GAT uses \ac{gat} convolutions~\citep{velickovic2018}.
Note that those three convolution operators are homogeneous, i.e., they assume an homophilic input graph.
This choice was made because most node classification datasets, including the chosen benchmark datasets, satisfy the homophily assumption.
The three \ac{cuqgnn} variants are compared against the following four baseline models:
The standard \textbf{\ac{gpn}} model using \ac{llop}~\citep{stadler2021}, \textbf{\ac{lopgpn}} which uses \ac{lop}~\citep{damke2024}, the first-order \textbf{\ac{appnp}} model~\citep{gasteiger2018} and the parameter-free \ac{gkde}~(\textbf{GKDE}) model~\citep{zhao2020}.
The hyperparameters and the training schedules for the models are chosen as described by \citet{stadler2021}.
The sizes and parameters of $h_{\mathit{enc}}$, $\mathbf{W}_{\mathit{lat}}$ and the \ac{postnet} model in \ac{cuqgnn}~(\cref{eq:cuqgnn}) are chosen as in the \ac{gpn} model~\citep{stadler2021}.
For CUQ-PPR, the \ac{appnp} convolution model is parameterized like the standalone \ac{appnp} model, i.e., $L = 10$ power iteration steps and a teleport probability of $\varepsilon = 0.1$~(\cref{eq:gpn-ppr}).
For CUQ-GCN and CUQ-GAT, the \ac{gnn} models consist of two convolution layers.
For \ac{gcn}, ReLU is used as activation function; for \ac{gat}, we use a single attention head with an \ac{elu} activation between the convolution layers.

\paragraph{Evaluation Metrics}
We assess the quality of the following five uncertainty estimates:
The entropy-based \ac{tu}, \ac{au} and \ac{eu} (\cref{eq:tu,eq:au,eq:eu}), the pseudo-count based $\mathrm{EU}_{\textrm{PC}}$ (\cref{sec:um:confidence}) and the second-order epistemic uncertainty $\mathrm{EU}_{\textrm{SO}}$ (\cref{eq:eu-so-entropy}).
Since \ac{appnp} only produces first-order predictions, we only evaluate the \ac{tu} of this model.
All other models predict second-order Dirichlet distributions and are evaluated using all four uncertainty estimates.

\subsection{Accuracy-Rejection Curves}\label{sec:eval:arc}

\newcommand*{\accrejPlot}[4]{\begin{tikzpicture}[inner frame sep=0]
    \small
    \begin{axis}[
        width=0.146\linewidth,
        height=1.46cm,
        scale only axis,
        xlabel={\ifthenelse{\equal{#1}{Arxiv}}{#4}{}},
        ylabel={\ifthenelse{\equal{#2}{sample_total_entropy}}{\scriptsize\ifthenelse{\equal{#1}{Computers}\or\equal{#1}{Photos}}{A.#1}{\ifthenelse{\equal{#1}{Arxiv}}{OGBN #1}{#1}}}{}},
        grid=major,
        ymin=#3, ymax=1,
        xmin=0, xmax=0.99,
        label style={font=\footnotesize},
        tick label style={font=\tiny},
        y tick label style={/pgf/number format/.cd, fixed, fixed zerofill, precision=2, /tikz/.cd},
        legend pos=outer north east,
        legend style={font=\tiny, draw=none},
        legend image post style={line width=2pt},
        legend image code/.code={
            \draw[mark repeat=2,mark phase=2]
            plot coordinates {
                (0cm,0cm)
                (0.08cm,0cm)        
                (0.16cm,0cm)         
            };%
        },
        inner frame sep=0,
    ]

    \addplot [color=black, line width=1pt] table [x=p, y=gpnMean, col sep=comma] {tables/acc_rej_#2_#1.csv};
    \addplot [color=t_gray, line width=1pt] table [x=p, y=gpnLOPMean, col sep=comma] {tables/acc_rej_#2_#1.csv};
    \addplot [color=t_green, line width=1pt] table [x=p, y=cuqPPRMean, col sep=comma] {tables/acc_rej_#2_#1.csv};
    \addplot [color=t_blue, line width=1pt] table [x=p, y=cuqGCNMean, col sep=comma] {tables/acc_rej_#2_#1.csv};
    \addplot [color=t_red, line width=1pt] table [x=p, y=cuqGATMean, col sep=comma] {tables/acc_rej_#2_#1.csv};

    \addSE{tables/acc_rej_#2_#1.csv}{gpn}{black}
    \addSE{tables/acc_rej_#2_#1.csv}{gpnLOP}{t_gray}
    \addSE{tables/acc_rej_#2_#1.csv}{cuqPPR}{t_green}
    \addSE{tables/acc_rej_#2_#1.csv}{cuqGCN}{t_blue}
    \addSE{tables/acc_rej_#2_#1.csv}{cuqGAT}{t_red}

    \ifthenelse{\equal{#2}{sample_epistemic}}{%
        \legend{GPN, LOP-GPN, CUQ-PPR, CUQ-GCN, CUQ-GAT}%
    }{}
    \end{axis}
\end{tikzpicture}}
\newcommand*{\accrejPlots}[2]{
    \hfill%
    \accrejPlot{#1}{sample_total_entropy}{#2}{$\mathrm{TU}$}%
    \hspace{-0.5em}\accrejPlot{#1}{sample_aleatoric_entropy}{#2}{$\mathrm{AU}$}%
    \hspace{-0.5em}\accrejPlot{#1}{sample_epistemic_entropy_diff}{#2}{$\mathrm{EU}$}%
    \hspace{-0.5em}\accrejPlot{#1}{sample_epistemic}{#2}{$\mathrm{EU}_{\mathrm{PC}}$}%
    \hfill
}
\begin{figure}[t!]
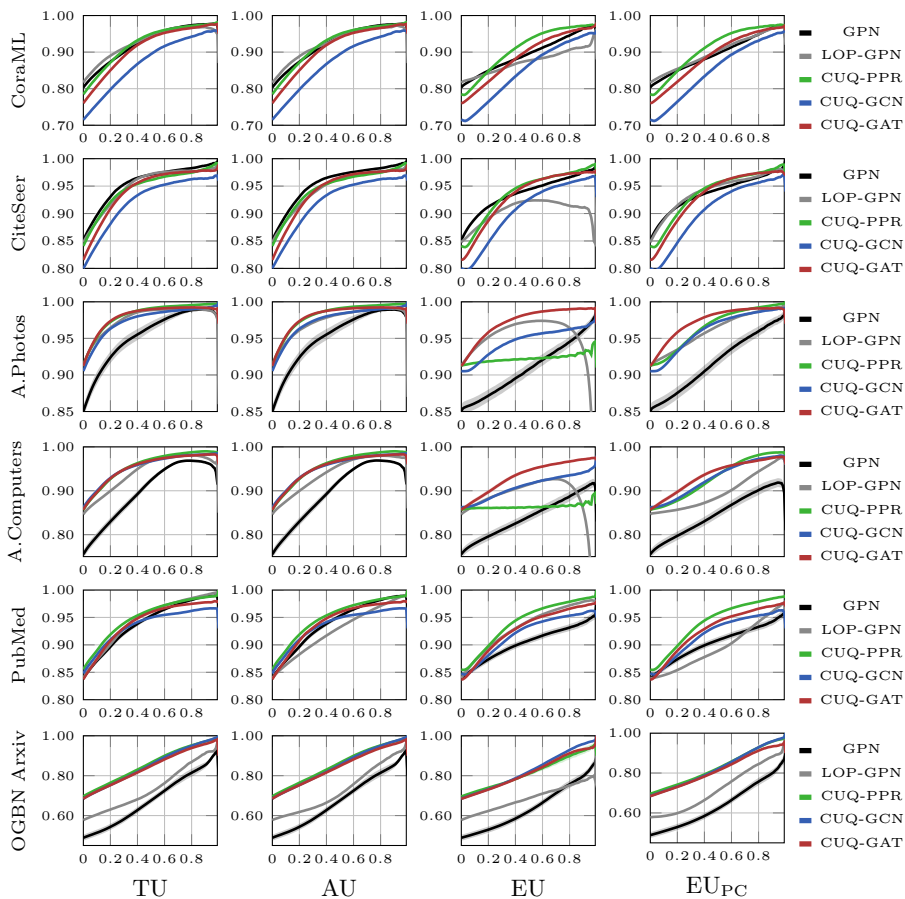

    \centering
    \accrejPlots{CoraML}{0.7}\\[-9pt]%
    \accrejPlots{CiteSeer}{0.8}\\[-9pt]%
    \accrejPlots{Photos}{0.85}\\[-9pt]%
    \accrejPlots{Computers}{0.75}\\[-9pt]%
    \accrejPlots{PubMed}{0.8}\\[-9pt]%
    \accrejPlots{Arxiv}{0.45}%
    \caption{%
        \Acl*{arc} for different uncertainty measures.
        The x-axis represents the fraction of rejected test instances; the y-axis represents the test accuracy for a given rejection rate.
        The (small) shaded areas behind the curves represent the estimate's standard error.
    }
    \label{fig:accrej}
\end{figure}
We begin with a comparison of \ac{cuqgnn} and \ac{gpn} using so-called \acfp{arc}.
The curves show the test accuracies of each model when discarding all test instances below a given uncertainty threshold.
For an uncertainty measure that captures predictive uncertainty well, the accuracy should monotonically increase with the rejection rate and, ideally, approach $100\%$.
\Cref{fig:accrej} shows the \acp{arc} of \ac{gpn} and the three evaluated \ac{cuqgnn} variants using the entropy-based \ac{tu} and \ac{eu} measures and the pseudo-count based $\mathrm{EU}_{\mathrm{PC}}$.

The accuracy of the \ac{cuqgnn} models when looking at the entire dataset, i.e., at $0\%$ rejection, is either close to or significantly higher than that of, both, \ac{gpn} and \ac{lopgpn}.
While \ac{cuqgnn} is motivated by the idea of behavioral pooling to resolve conflicts between nodes more flexibly and thereby improve its uncertainty estimates, the increased flexibility also improves the model's generalization performance in general.
Within the three evaluated types of \ac{cuqgnn} models, there is no clear winner across all datasets, supporting the idea that the appropriateness of a given pooling mechanism is highly domain-dependent.

As should be expected, accuracy increases monotonically with increasing rejection rates across all uncertainty measures, models, and datasets.
Additionally, all \ac{cuqgnn} variants reach accuracies close to $100\%$ as the rejection rate increases.
For \ac{gpn}, on the other hand, the delta to $100\%$ at high rejection rates is significantly higher.
There are two notable exceptions to those observations:
First, the \acp{arc} of CUQ-PPR are nearly flat on the Amazon Photos and Amazon Computers datasets when using entropy-based \ac{eu}.
Using pseudo-count-based \acp{eu}, the \ac{arc} of CUQ-PPR on those datasets is, however, increasing.
This discrepancy illustrates the previously mentioned problem that entropy-based measures are not always appropriate measures of uncertainty~\citep{wimmer2023}.
Second, using the entropy-based \ac{tu}, \ac{au} and \ac{eu} measures, the \acp{arc} of \ac{gpn} and \ac{lopgpn} go down for high rejection rates on the two Amazon datasets, indicating that the models incorrectly assign high confidences to wrong predictions.

To summarize, \cref{fig:accrej} shows that \textbf{\ac{cuqgnn} achieves a similar or better predictive performance than \ac{gpn} and \ac{lopgpn}}.
Additionally, the \textbf{uncertainty estimates of \ac{cuqgnn} behave better than those of \ac{gpn} and \ac{lopgpn}}, in the sense that they lead to near-perfect accuracies at high rejection rates.

\subsection{OOD Detection}\label{sec:eval:ood}

\begin{table}[t!]
    \caption{OOD detection performance of OOD vs ID vertices and ID accuracies.}
    \label{tbl:ood}
    \centering
    {\tiny%
    \begin{filecontents*}{tables/id_ood.csv}
id,dataset,model,acc,accSE,accBest,oodLocIdAcc,oodLocIdAccSE,oodLocIdAccBest,oodLocOodAcc,oodLocOodAccSE,oodLocOodAccBest,oodLocOodTotal,oodLocOodTotalSE,oodLocOodTotalBest,oodLocOodTotalEntropy,oodLocOodTotalEntropySE,oodLocOodTotalEntropyBest,oodLocOodAleatoric,oodLocOodAleatoricSE,oodLocOodAleatoricBest,oodLocOodAleatoricEntropy,oodLocOodAleatoricEntropySE,oodLocOodAleatoricEntropyBest,oodLocOodEpistemic,oodLocOodEpistemicSE,oodLocOodEpistemicBest,oodLocOodEpistemicEntropy,oodLocOodEpistemicEntropySE,oodLocOodEpistemicEntropyBest,oodLocOodEpistemicEntropyDiff,oodLocOodEpistemicEntropyDiffSE,oodLocOodEpistemicEntropyDiffBest,oodLocTotalEntropyChange,oodLocAleatoricEntropyChange,oodLocEpistemicEntropyChange,oodLocEpistemicEntropyDiffChange,oodNormalIdAcc,oodNormalIdAccSE,oodNormalIdAccBest,oodNormalOodAcc,oodNormalOodAccSE,oodNormalOodAccBest,oodNormalOodTotal,oodNormalOodTotalSE,oodNormalOodTotalBest,oodNormalOodTotalEntropy,oodNormalOodTotalEntropySE,oodNormalOodTotalEntropyBest,oodNormalOodAleatoric,oodNormalOodAleatoricSE,oodNormalOodAleatoricBest,oodNormalOodAleatoricEntropy,oodNormalOodAleatoricEntropySE,oodNormalOodAleatoricEntropyBest,oodNormalOodEpistemic,oodNormalOodEpistemicSE,oodNormalOodEpistemicBest,oodNormalOodEpistemicEntropy,oodNormalOodEpistemicEntropySE,oodNormalOodEpistemicEntropyBest,oodNormalOodEpistemicEntropyDiff,oodNormalOodEpistemicEntropyDiffSE,oodNormalOodEpistemicEntropyDiffBest,oodNormalTotalEntropyChange,oodNormalAleatoricEntropyChange,oodNormalEpistemicEntropyChange,oodNormalEpistemicEntropyDiffChange,oodBerIdAcc,oodBerIdAccSE,oodBerIdAccBest,oodBerOodAcc,oodBerOodAccSE,oodBerOodAccBest,oodBerOodTotal,oodBerOodTotalSE,oodBerOodTotalBest,oodBerOodTotalEntropy,oodBerOodTotalEntropySE,oodBerOodTotalEntropyBest,oodBerOodAleatoric,oodBerOodAleatoricSE,oodBerOodAleatoricBest,oodBerOodAleatoricEntropy,oodBerOodAleatoricEntropySE,oodBerOodAleatoricEntropyBest,oodBerOodEpistemic,oodBerOodEpistemicSE,oodBerOodEpistemicBest,oodBerOodEpistemicEntropy,oodBerOodEpistemicEntropySE,oodBerOodEpistemicEntropyBest,oodBerOodEpistemicEntropyDiff,oodBerOodEpistemicEntropyDiffSE,oodBerOodEpistemicEntropyDiffBest,oodBerTotalEntropyChange,oodBerAleatoricEntropyChange,oodBerEpistemicEntropyChange,oodBerEpistemicEntropyDiffChange
0,CoraML,APPNP,84.35,0.00,1,90.44,0.00,1,0.00,0.00,1,86.40,0.00,1,87.45,0.00,1,86.40,0.00,1,,0.00,0,,0.00,0,,0.00,0,,0.00,0,+125.64,+125.64,,,43.62,0.00,0,18.69,0.00,0,14.48,0.00,0,13.26,0.00,0,14.48,0.00,0,,0.00,0,,0.00,0,,0.00,0,,0.00,0,-84.97,-84.97,,,84.14,0.00,1,80.60,0.00,1,63.85,0.00,1,65.02,0.00,1,63.85,0.00,1,,0.00,0,,0.00,0,,0.00,0,,0.00,0,,,,
1,CoraML,GKDE,71.73,0.00,0,83.01,0.00,0,0.00,0.00,1,77.65,0.00,0,77.21,0.00,0,77.65,0.00,0,35.35,0.00,0,74.00,0.00,0,76.46,0.00,0,69.16,0.00,0,+0.400,-0.401,-198.1,+3.426,71.96,0.00,0,69.69,0.00,0,48.83,0.00,0,48.71,0.00,0,48.83,0.00,0,50.76,0.00,0,48.45,0.00,0,48.68,0.00,0,49.19,0.00,0,-0.003,+0.017,-200.0,-0.100,71.96,0.00,0,69.69,0.00,0,48.83,0.00,0,48.71,0.00,0,48.83,0.00,0,50.76,0.00,0,48.45,0.00,0,48.68,0.00,0,49.19,0.00,0,,,-200.0,-0.100
2,CoraML,GPN,80.34,0.00,0,89.36,0.00,0,0.00,0.00,1,84.29,0.00,0,85.51,0.00,0,84.29,0.00,0,85.49,0.00,0,87.11,0.00,1,89.15,0.00,1,86.23,0.00,1,+104.49,+104.42,-174.9,+229.74,17.86,0.00,0,18.19,0.00,0,89.14,0.00,0,96.59,0.00,1,89.14,0.00,0,96.80,0.00,1,70.62,0.00,0,75.66,0.00,0,70.89,0.00,0,+2.038,+2.018,-196.9,+24.074,80.36,0.00,0,78.41,0.00,0,54.17,0.00,0,54.44,0.00,0,54.17,0.00,0,54.44,0.00,0,54.17,0.00,0,87.19,0.00,1,51.56,0.00,0,,,-176.3,+15.300
3,CoraML,LOP-GPN,81.69,0.00,0,89.34,0.00,0,0.00,0.00,1,84.45,0.00,0,85.67,0.00,0,84.45,0.00,0,88.51,0.00,1,84.72,0.00,0,79.26,0.00,0,45.18,0.00,0,+56.987,+81.573,-174.8,-5.327,81.74,0.00,1,79.92,0.00,1,65.19,0.00,0,69.50,0.00,0,65.19,0.00,0,61.06,0.00,0,60.15,0.00,0,83.69,0.00,0,82.08,0.00,0,+20.342,+15.113,-180.7,+33.451,81.29,0.00,0,76.82,0.00,0,62.04,0.00,0,62.20,0.00,0,62.04,0.00,0,59.69,0.00,0,54.40,0.00,0,56.74,0.00,0,64.31,0.00,1,,,-195.8,+16.870
4,CoraML,CUQ-PPR,78.48,0.00,0,87.81,0.00,0,0.00,0.00,1,80.51,0.00,0,81.18,0.00,0,80.51,0.00,0,81.20,0.00,0,78.23,0.00,0,80.37,0.00,0,78.10,0.00,0,+112.09,+112.21,-177.0,+35.971,26.85,0.00,0,12.16,0.00,0,91.45,0.00,1,85.28,0.00,0,91.45,0.00,1,53.09,0.00,0,91.24,0.00,1,88.23,0.00,1,89.97,0.00,1,+27.960,+15.753,-137.6,+146.20,77.67,0.00,0,69.64,0.00,0,60.42,0.00,0,60.82,0.00,0,60.42,0.00,0,60.82,0.00,1,59.41,0.00,1,60.06,0.00,0,59.38,0.00,0,+29.184,+29.195,-193.3,+2.555
5,CoraML,CUQ-GCN,71.52,0.00,0,82.42,0.00,0,0.00,0.00,1,73.69,0.00,0,74.41,0.00,0,73.69,0.00,0,74.44,0.00,0,69.27,0.00,0,72.47,0.00,0,69.23,0.00,0,+63.063,+63.259,-184.6,-29.97,27.47,0.00,0,12.45,0.00,0,80.45,0.00,0,75.03,0.00,0,80.45,0.00,0,57.37,0.00,0,80.28,0.00,0,77.57,0.00,0,78.97,0.00,0,+25.966,+18.444,-140.4,+78.834,70.11,0.00,0,61.94,0.00,0,57.20,0.00,0,57.34,0.00,0,57.20,0.00,0,57.34,0.00,0,55.23,0.00,0,56.27,0.00,0,55.22,0.00,0,+15.048,+15.082,-196.2,-22.15
6,CoraML,CUQ-GAT,76.02,0.00,0,85.97,0.00,0,0.00,0.00,1,79.35,0.00,0,79.88,0.00,0,79.35,0.00,0,79.88,0.00,0,78.10,0.00,0,79.86,0.00,0,78.10,0.00,0,+96.106,+96.109,-177.9,+92.141,30.56,0.00,0,14.15,0.00,0,71.17,0.00,0,69.76,0.00,0,71.17,0.00,0,60.79,0.00,0,71.40,0.00,0,70.66,0.00,0,71.10,0.00,0,+22.820,+17.063,-146.0,+61.155,74.71,0.00,0,68.91,0.00,0,57.31,0.00,0,57.37,0.00,0,57.31,0.00,0,57.37,0.00,0,55.17,0.00,0,56.45,0.00,0,55.15,0.00,0,+15.747,+15.754,-196.3,-1.722
7,CiteSeer,APPNP,86.04,0.00,1,87.75,0.00,1,0.00,0.00,1,85.33,0.00,1,85.80,0.00,1,85.33,0.00,1,,0.00,0,,0.00,0,,0.00,0,,0.00,0,+159.54,+159.54,,,72.72,0.00,0,25.05,0.00,0,26.62,0.00,0,23.76,0.00,0,26.62,0.00,0,,0.00,0,,0.00,0,,0.00,0,,0.00,0,-62.75,-62.75,,,85.65,0.00,1,83.64,0.00,1,68.65,0.00,0,70.25,0.00,0,68.65,0.00,0,,0.00,0,,0.00,0,,0.00,0,,0.00,0,,,,
8,CiteSeer,GKDE,65.85,0.00,0,73.73,0.00,0,0.00,0.00,1,81.33,0.00,0,77.89,0.00,0,81.33,0.00,0,37.03,0.00,0,81.30,0.00,1,79.44,0.00,0,65.08,0.00,0,+0.131,-0.291,-199.3,+1.673,65.79,0.00,0,66.45,0.00,0,50.18,0.00,0,50.14,0.00,0,50.18,0.00,0,51.33,0.00,0,50.17,0.00,0,50.04,0.00,0,48.80,0.00,0,+0.002,+0.019,-199.9,-0.070,65.79,0.00,0,66.45,0.00,0,50.18,0.00,0,50.14,0.00,0,50.18,0.00,0,51.33,0.00,0,50.17,0.00,0,50.04,0.00,0,48.80,0.00,0,,,-199.9,-0.070
9,CiteSeer,GPN,85.13,0.00,0,87.23,0.00,0,0.00,0.00,1,80.87,0.00,0,81.53,0.00,0,80.87,0.00,0,81.52,0.00,1,76.76,0.00,0,80.62,0.00,1,76.78,0.00,0,+98.306,+98.324,-181.8,+84.116,17.45,0.00,0,17.30,0.00,0,86.96,0.00,0,93.00,0.00,0,86.96,0.00,0,93.22,0.00,1,79.91,0.00,0,86.34,0.00,0,80.32,0.00,0,+5.183,+5.158,-195.1,+8.989,84.82,0.00,0,82.24,0.00,0,61.82,0.00,0,62.58,0.00,0,61.82,0.00,0,62.58,0.00,0,60.16,0.00,0,91.73,0.00,1,53.65,0.00,0,,,-171.7,+32.222
10,CiteSeer,LOP-GPN,84.74,0.00,0,87.16,0.00,0,0.00,0.00,1,81.66,0.00,0,82.19,0.00,0,81.66,0.00,0,80.57,0.00,0,75.48,0.00,0,73.12,0.00,0,73.09,0.00,0,+80.393,+77.788,-185.4,+84.756,84.41,0.00,1,82.74,0.00,1,75.82,0.00,0,81.65,0.00,0,75.82,0.00,0,76.92,0.00,0,59.20,0.00,0,86.81,0.00,0,81.26,0.00,0,+55.932,+54.267,-172.5,+58.547,84.31,0.00,0,81.37,0.00,0,75.14,0.00,1,78.74,0.00,1,75.14,0.00,1,78.43,0.00,1,58.12,0.00,0,68.98,0.00,0,68.54,0.00,1,,,-189.9,+33.167
11,CiteSeer,CUQ-PPR,84.09,0.00,0,86.41,0.00,0,0.00,0.00,1,79.64,0.00,0,79.39,0.00,0,79.64,0.00,0,79.42,0.00,0,76.34,0.00,0,77.50,0.00,0,76.34,0.00,0,+85.847,+85.964,-180.7,+37.981,60.71,0.00,0,20.55,0.00,0,94.28,0.00,1,94.61,0.00,1,94.28,0.00,1,78.28,0.00,0,94.84,0.00,1,94.71,0.00,1,94.82,0.00,1,+66.721,+44.579,-126.2,+744.13,83.40,0.00,0,77.96,0.00,0,67.00,0.00,0,67.33,0.00,0,67.00,0.00,0,67.34,0.00,0,62.90,0.00,1,65.42,0.00,0,62.90,0.00,0,+43.179,+43.195,-191.3,+4.230
12,CiteSeer,CUQ-GCN,80.03,0.00,0,82.63,0.00,0,0.00,0.00,1,72.40,0.00,0,72.32,0.00,0,72.40,0.00,0,72.48,0.00,0,67.15,0.00,0,69.78,0.00,0,67.17,0.00,0,+57.354,+58.954,-185.3,-55.15,52.64,0.00,0,18.08,0.00,0,90.21,0.00,0,90.50,0.00,0,90.21,0.00,0,79.52,0.00,0,91.30,0.00,0,90.82,0.00,0,91.25,0.00,0,+76.059,+55.453,-129.0,+445.83,78.91,0.00,0,69.88,0.00,0,64.39,0.00,0,64.47,0.00,0,64.39,0.00,0,64.49,0.00,0,59.99,0.00,0,62.24,0.00,0,59.95,0.00,0,+36.203,+36.281,-192.3,-24.33
13,CiteSeer,CUQ-GAT,81.55,0.00,0,83.00,0.00,0,0.00,0.00,1,78.77,0.00,0,79.10,0.00,0,78.77,0.00,0,80.11,0.00,0,77.48,0.00,0,78.43,0.00,0,77.48,0.00,1,+94.610,+96.784,-175.5,-6.349,54.77,0.00,0,20.05,0.00,0,85.62,0.00,0,86.00,0.00,0,85.62,0.00,0,77.26,0.00,0,87.12,0.00,0,86.81,0.00,0,87.10,0.00,0,+70.076,+54.018,-132.3,+302.71,80.54,0.00,0,74.39,0.00,0,64.20,0.00,0,64.67,0.00,0,64.20,0.00,0,64.68,0.00,0,60.35,0.00,0,63.13,0.00,0,60.24,0.00,0,+41.497,+41.532,-191.7,-0.388
14,Amazon\\Photos,APPNP,91.68,0.00,1,94.99,0.00,1,0.00,0.00,1,75.58,0.00,0,75.12,0.00,0,75.58,0.00,0,,0.00,0,,0.00,0,,0.00,0,,0.00,0,+79.465,+79.465,,,40.44,0.00,0,19.84,0.00,0,14.42,0.00,0,13.17,0.00,0,14.42,0.00,0,,0.00,0,,0.00,0,,0.00,0,,0.00,0,-85.22,-85.22,,,91.83,0.00,1,86.76,0.00,0,63.58,0.00,0,63.55,0.00,0,63.58,0.00,0,,0.00,0,,0.00,0,,0.00,0,,0.00,0,,,,
15,Amazon\\Photos,GKDE,76.16,0.00,0,85.45,0.00,0,0.00,0.00,1,75.92,0.00,0,70.20,0.00,0,75.92,0.00,0,55.45,0.00,0,60.80,0.00,0,66.83,0.00,0,61.23,0.00,0,+1.361,+0.435,-196.3,+5.897,76.19,0.00,0,75.87,0.00,0,49.21,0.00,0,49.07,0.00,0,49.21,0.00,0,50.74,0.00,0,48.87,0.00,0,48.95,0.00,0,48.76,0.00,0,-0.064,+0.047,-200.2,-0.729,76.19,0.00,0,75.87,0.00,0,49.21,0.00,0,49.07,0.00,0,49.21,0.00,0,50.74,0.00,0,48.87,0.00,0,48.95,0.00,0,48.76,0.00,0,,,-200.2,-0.729
16,Amazon\\Photos,GPN,85.02,0.00,0,91.49,0.00,0,0.00,0.00,1,75.39,0.00,0,76.29,0.00,0,75.39,0.00,0,76.29,0.00,0,87.50,0.00,1,86.05,0.00,1,86.54,0.00,1,+57.315,+57.311,-184.5,+196.11,12.63,0.00,0,12.83,0.00,0,95.40,0.00,1,89.43,0.00,0,95.40,0.00,1,91.07,0.00,1,59.86,0.00,0,63.56,0.00,0,60.06,0.00,0,+1.613,+1.579,-197.4,+17.020,84.79,0.00,0,81.95,0.00,0,54.36,0.00,0,54.55,0.00,0,54.36,0.00,0,54.55,0.00,0,54.29,0.00,0,80.17,0.00,1,49.87,0.00,0,,,-179.2,+6.485
17,Amazon\\Photos,LOP-GPN,91.19,0.00,0,94.00,0.00,0,0.00,0.00,1,83.86,0.00,1,86.50,0.00,1,83.86,0.00,1,83.88,0.00,1,76.63,0.00,0,83.87,0.00,0,80.02,0.00,0,+113.05,+111.07,-184.7,+116.42,91.18,0.00,1,86.89,0.00,1,84.54,0.00,0,91.76,0.00,1,84.54,0.00,0,87.29,0.00,0,61.15,0.00,0,95.74,0.00,1,89.36,0.00,0,+93.379,+98.786,-165.4,+85.072,91.10,0.00,0,77.69,0.00,0,79.78,0.00,1,79.21,0.00,1,79.78,0.00,1,70.67,0.00,1,53.39,0.00,0,61.00,0.00,0,78.59,0.00,1,,,-195.6,+69.456
18,Amazon\\Photos,CUQ-PPR,91.29,0.00,0,92.76,0.00,0,0.00,0.00,1,75.00,0.00,0,75.18,0.00,0,75.00,0.00,0,75.19,0.00,0,69.96,0.00,0,72.69,0.00,0,65.17,0.00,0,+148.63,+148.72,-184.7,-63.91,20.75,0.00,0,8.12,0.00,0,93.57,0.00,0,89.93,0.00,0,93.57,0.00,0,58.20,0.00,0,93.37,0.00,1,89.38,0.00,0,93.41,0.00,1,+36.073,+22.734,-131.6,+192.64,90.53,0.00,0,79.35,0.00,0,62.98,0.00,0,63.08,0.00,0,62.98,0.00,0,63.08,0.00,0,59.03,0.00,1,61.32,0.00,0,50.10,0.00,0,+80.089,+80.104,-194.5,-66.99
19,Amazon\\Photos,CUQ-GCN,90.53,0.00,0,92.26,0.00,0,0.00,0.00,1,76.08,0.00,0,76.11,0.00,0,76.08,0.00,0,76.12,0.00,0,69.81,0.00,0,72.58,0.00,0,69.44,0.00,0,+152.40,+153.13,-182.2,-59.33,15.74,0.00,0,9.47,0.00,0,66.05,0.00,0,62.53,0.00,0,66.05,0.00,0,50.99,0.00,0,65.80,0.00,0,61.89,0.00,0,65.94,0.00,0,+11.817,+9.843,-160.3,+22.835,90.00,0.00,0,87.01,0.00,0,52.95,0.00,0,52.97,0.00,0,52.95,0.00,0,52.97,0.00,0,51.81,0.00,0,52.39,0.00,0,51.39,0.00,0,+17.981,+18.061,-198.6,-55.67
20,Amazon\\Photos,CUQ-GAT,91.25,0.00,0,92.69,0.00,0,0.00,0.00,1,78.31,0.00,0,78.82,0.00,0,78.31,0.00,0,78.82,0.00,0,76.87,0.00,0,78.13,0.00,0,76.79,0.00,0,+172.35,+172.36,-179.8,+111.92,23.00,0.00,0,8.43,0.00,0,62.60,0.00,0,62.11,0.00,0,62.60,0.00,0,59.52,0.00,0,62.65,0.00,0,62.07,0.00,0,62.66,0.00,0,+20.067,+18.136,-151.3,+30.545,90.59,0.00,0,88.56,0.00,1,53.31,0.00,0,53.33,0.00,0,53.31,0.00,0,53.33,0.00,0,52.16,0.00,0,52.71,0.00,0,52.09,0.00,0,+14.015,+14.035,-198.2,-76.15
21,Amazon\\Computers,APPNP,80.92,0.00,0,87.99,0.00,0,0.00,0.00,1,80.31,0.00,0,79.32,0.00,0,80.31,0.00,0,,0.00,0,,0.00,0,,0.00,0,,0.00,0,+76.515,+76.515,,,42.81,0.00,0,21.64,0.00,0,16.93,0.00,0,15.58,0.00,0,16.93,0.00,0,,0.00,0,,0.00,0,,0.00,0,,0.00,0,-82.43,-82.43,,,81.11,0.00,0,75.97,0.00,0,64.20,0.00,0,64.62,0.00,0,64.20,0.00,0,,0.00,0,,0.00,0,,0.00,0,,0.00,0,,,,
22,Amazon\\Computers,GKDE,64.04,0.00,0,71.26,0.00,0,0.00,0.00,1,75.20,0.00,0,76.38,0.00,0,75.20,0.00,0,70.52,0.00,0,74.37,0.00,0,76.20,0.00,0,74.46,0.00,0,+9.293,+5.394,-180.5,+35.198,64.01,0.00,0,64.34,0.00,0,49.88,0.00,0,49.92,0.00,0,49.88,0.00,0,49.81,0.00,0,50.09,0.00,0,49.99,0.00,0,50.03,0.00,0,-0.027,-0.034,-200.0,+0.026,64.01,0.00,0,64.34,0.00,0,49.88,0.00,0,49.92,0.00,0,49.88,0.00,0,49.81,0.00,0,50.09,0.00,0,49.99,0.00,0,50.03,0.00,0,,,-200.0,+0.026
23,Amazon\\Computers,GPN,75.53,0.00,0,82.17,0.00,0,0.00,0.00,1,77.17,0.00,0,79.17,0.00,0,77.17,0.00,0,79.17,0.00,0,81.01,0.00,1,83.77,0.00,0,76.65,0.00,1,+44.150,+44.151,-189.0,+15.961,16.39,0.00,0,17.03,0.00,0,83.24,0.00,0,88.69,0.00,0,83.24,0.00,0,89.62,0.00,0,62.12,0.00,0,65.90,0.00,0,62.32,0.00,0,+2.973,+2.919,-197.0,+19.750,75.71,0.00,0,74.51,0.00,0,54.28,0.00,0,54.67,0.00,0,54.28,0.00,0,54.67,0.00,0,56.46,0.00,0,85.57,0.00,1,51.14,0.00,0,,,-174.9,+9.359
24,Amazon\\Computers,LOP-GPN,84.62,0.00,0,90.28,0.00,0,0.00,0.00,1,83.24,0.00,1,85.00,0.00,1,83.24,0.00,1,88.08,0.00,1,77.98,0.00,0,83.80,0.00,1,68.40,0.00,0,+85.739,+105.73,-189.0,+48.044,84.49,0.00,1,81.42,0.00,1,86.47,0.00,0,94.76,0.00,1,86.47,0.00,0,90.93,0.00,1,65.30,0.00,0,98.15,0.00,1,88.33,0.00,0,+84.477,+87.874,-162.6,+78.112,84.35,0.00,0,67.95,0.00,0,79.62,0.00,1,79.78,0.00,1,79.62,0.00,1,72.26,0.00,1,51.59,0.00,0,61.91,0.00,0,75.86,0.00,1,,,-197.2,+59.183
25,Amazon\\Computers,CUQ-PPR,85.61,0.00,0,90.53,0.00,0,0.00,0.00,1,72.91,0.00,0,72.78,0.00,0,72.91,0.00,0,72.78,0.00,0,61.73,0.00,0,66.21,0.00,0,50.65,0.00,0,+120.25,+120.24,-191.6,+464.28,24.02,0.00,0,7.68,0.00,0,90.53,0.00,1,86.02,0.00,0,90.53,0.00,1,41.73,0.00,0,90.44,0.00,1,84.00,0.00,0,89.86,0.00,1,+17.556,+8.681,-144.3,+107.98,84.71,0.00,0,74.79,0.00,0,67.50,0.00,0,68.26,0.00,0,67.50,0.00,0,68.26,0.00,0,64.27,0.00,1,66.91,0.00,0,49.98,0.00,0,+70.944,+70.950,-193.2,-97.02
26,Amazon\\Computers,CUQ-GCN,86.19,0.00,1,91.87,0.00,1,0.00,0.00,1,68.71,0.00,0,68.34,0.00,0,68.71,0.00,0,68.34,0.00,0,60.06,0.00,0,62.51,0.00,0,59.46,0.00,0,+104.49,+104.73,-191.0,-31.45,13.06,0.00,0,7.68,0.00,0,63.28,0.00,0,60.39,0.00,0,63.28,0.00,0,49.91,0.00,0,62.94,0.00,0,58.91,0.00,0,62.43,0.00,0,+8.011,+6.671,-169.5,+15.723,85.63,0.00,1,83.07,0.00,1,53.95,0.00,0,54.14,0.00,0,53.95,0.00,0,54.14,0.00,0,52.61,0.00,0,53.51,0.00,0,51.75,0.00,0,+16.435,+16.537,-198.2,-71.44
27,Amazon\\Computers,CUQ-GAT,85.66,0.00,0,91.81,0.00,0,0.00,0.00,1,79.57,0.00,0,79.74,0.00,0,79.57,0.00,0,79.78,0.00,0,75.57,0.00,0,76.91,0.00,0,75.31,0.00,0,+187.91,+189.00,-180.3,-11.00,16.14,0.00,0,5.89,0.00,0,58.43,0.00,0,58.25,0.00,0,58.43,0.00,0,57.18,0.00,0,58.44,0.00,0,58.16,0.00,0,58.41,0.00,0,+13.430,+12.367,-161.6,+19.408,85.03,0.00,0,83.01,0.00,0,53.98,0.00,0,54.08,0.00,0,53.98,0.00,0,54.09,0.00,0,52.85,0.00,0,53.37,0.00,0,52.77,0.00,0,+13.384,+13.606,-198.2,-29.25
28,PubMed,APPNP,86.63,0.00,1,94.59,0.00,1,0.00,0.00,1,69.38,0.00,0,69.38,0.00,0,69.38,0.00,0,,0.00,0,,0.00,0,,0.00,0,,0.00,0,+84.405,+84.405,,,59.59,0.00,0,37.81,0.00,0,12.47,0.00,0,10.95,0.00,0,12.47,0.00,0,,0.00,0,,0.00,0,,0.00,0,,0.00,0,-86.78,-86.78,,,86.73,0.00,1,82.75,0.00,1,64.97,0.00,0,66.90,0.00,0,64.97,0.00,0,,0.00,0,,0.00,0,,0.00,0,,0.00,0,,,,
29,PubMed,GKDE,76.99,0.00,0,87.93,0.00,0,0.00,0.00,1,71.50,0.00,1,71.51,0.00,1,71.50,0.00,1,39.69,0.00,0,69.74,0.00,0,71.00,0.00,0,65.30,0.00,0,+0.774,-0.536,-120.3,+4.457,76.96,0.00,0,77.23,0.00,0,49.94,0.00,0,49.94,0.00,0,49.94,0.00,0,50.16,0.00,0,49.87,0.00,0,49.93,0.00,0,49.79,0.00,0,-0.014,-0.002,-200.0,-0.057,76.96,0.00,0,77.23,0.00,0,49.94,0.00,0,49.94,0.00,0,49.94,0.00,0,50.16,0.00,0,49.87,0.00,0,49.93,0.00,0,49.79,0.00,0,,,-200.0,-0.057
30,PubMed,GPN,83.85,0.00,0,93.76,0.00,0,0.00,0.00,1,69.87,0.00,0,69.87,0.00,0,69.87,0.00,0,69.85,0.00,1,72.93,0.00,1,73.62,0.00,1,72.95,0.00,1,+57.817,+57.809,-185.8,+64.653,30.29,0.00,0,30.44,0.00,0,93.44,0.00,1,95.74,0.00,1,93.44,0.00,1,96.41,0.00,1,70.29,0.00,0,76.44,0.00,0,70.52,0.00,0,+4.252,+4.236,-196.3,+6.699,83.89,0.00,0,81.16,0.00,0,58.10,0.00,0,58.49,0.00,0,58.10,0.00,0,58.49,0.00,0,60.66,0.00,0,81.42,0.00,1,51.12,0.00,0,,,-175.0,+15.346
31,PubMed,LOP-GPN,83.94,0.00,0,93.23,0.00,0,0.00,0.00,1,68.96,0.00,0,68.96,0.00,0,68.96,0.00,0,69.10,0.00,0,64.87,0.00,0,64.30,0.00,0,66.28,0.00,0,+48.296,+44.303,-183.5,+56.243,83.99,0.00,1,81.68,0.00,1,77.03,0.00,0,82.71,0.00,0,77.03,0.00,0,75.93,0.00,0,67.74,0.00,0,91.84,0.00,0,80.37,0.00,0,+45.693,+37.657,-160.4,+60.553,83.52,0.00,0,74.83,0.00,0,70.26,0.00,1,70.46,0.00,1,70.26,0.00,1,67.93,0.00,1,55.83,0.00,0,60.86,0.00,0,65.49,0.00,1,,,-195.2,+36.812
32,PubMed,CUQ-PPR,85.50,0.00,0,94.28,0.00,0,0.00,0.00,1,65.19,0.00,0,65.19,0.00,0,65.19,0.00,0,65.19,0.00,0,64.15,0.00,0,65.22,0.00,0,64.19,0.00,0,+63.216,+63.339,-187.5,+37.281,44.25,0.00,0,27.92,0.00,0,92.74,0.00,0,91.93,0.00,0,92.74,0.00,0,59.89,0.00,0,93.45,0.00,1,92.97,0.00,1,93.18,0.00,1,+44.459,+20.391,-116.7,+404.71,85.51,0.00,0,79.46,0.00,0,62.83,0.00,0,63.01,0.00,0,62.83,0.00,0,63.01,0.00,0,61.05,0.00,1,62.39,0.00,0,61.04,0.00,0,+35.608,+35.622,-191.8,+3.799
33,PubMed,CUQ-GCN,84.79,0.00,0,94.03,0.00,0,0.00,0.00,1,61.81,0.00,0,61.88,0.00,0,61.81,0.00,0,62.35,0.00,0,59.62,0.00,0,61.09,0.00,0,59.61,0.00,0,+35.480,+39.497,-190.5,-10.96,45.53,0.00,0,26.63,0.00,0,83.48,0.00,0,81.47,0.00,0,83.48,0.00,0,61.04,0.00,0,83.69,0.00,0,82.65,0.00,0,83.07,0.00,0,+40.819,+26.373,-125.0,+138.38,84.67,0.00,0,76.50,0.00,0,62.62,0.00,0,62.70,0.00,0,62.62,0.00,0,62.71,0.00,0,58.94,0.00,0,61.04,0.00,0,58.92,0.00,0,+30.748,+30.829,-193.8,-18.00
34,PubMed,CUQ-GAT,83.72,0.00,0,92.86,0.00,0,0.00,0.00,1,63.16,0.00,0,63.32,0.00,0,63.16,0.00,0,64.40,0.00,0,59.07,0.00,0,61.98,0.00,0,59.09,0.00,0,+31.285,+36.690,-188.5,-5.401,46.02,0.00,0,28.35,0.00,0,70.53,0.00,0,69.93,0.00,0,70.53,0.00,0,61.10,0.00,0,71.06,0.00,0,70.66,0.00,0,70.81,0.00,0,+28.437,+20.348,-134.8,+70.363,83.50,0.00,0,78.47,0.00,0,58.82,0.00,0,58.96,0.00,0,58.82,0.00,0,59.10,0.00,0,55.30,0.00,0,57.60,0.00,0,55.30,0.00,0,+21.022,+21.652,-195.5,-28.52
35,OGBN\\Arxiv,APPNP,68.76,0.00,0,73.33,0.00,0,0.00,0.00,1,64.39,0.00,0,64.47,0.00,0,64.39,0.00,0,,0.00,0,,0.00,0,,0.00,0,,0.00,0,+28.927,+28.927,,,66.53,0.00,1,27.28,0.00,0,46.98,0.00,0,42.63,0.00,0,46.98,0.00,0,,0.00,0,,0.00,0,,0.00,0,,0.00,0,-12.93,-12.93,,,68.22,0.00,0,65.73,0.00,0,63.98,0.00,0,66.97,0.00,0,63.98,0.00,0,,0.00,0,,0.00,0,,0.00,0,,0.00,0,,,,
36,OGBN\\Arxiv,GKDE,56.97,0.00,0,60.04,0.00,0,0.00,0.00,1,69.41,0.00,1,68.69,0.00,0,69.41,0.00,1,66.97,0.00,0,67.46,0.00,0,68.14,0.00,0,66.70,0.00,0,+17.791,+16.112,-186.8,+34.402,56.91,0.00,0,57.59,0.00,1,49.63,0.00,0,49.52,0.00,0,49.63,0.00,0,49.55,0.00,0,49.47,0.00,0,49.46,0.00,0,49.48,0.00,0,-0.431,-0.385,-200.3,-0.939,56.91,0.00,0,57.59,0.00,0,49.63,0.00,0,49.52,0.00,0,49.63,0.00,0,49.55,0.00,0,49.47,0.00,0,49.46,0.00,0,49.48,0.00,0,,,-200.3,-0.939
37,OGBN\\Arxiv,GPN,48.87,0.00,0,55.36,0.00,0,0.00,0.00,1,64.31,0.00,0,67.08,0.00,0,64.31,0.00,0,67.08,0.00,0,66.91,0.00,0,68.79,0.00,0,66.92,0.00,0,+20.547,+20.544,-194.4,+30.225,4.35,0.00,0,4.38,0.00,0,67.63,0.00,0,77.08,0.00,0,67.63,0.00,0,77.45,0.00,0,66.92,0.00,0,69.93,0.00,0,67.14,0.00,0,+2.084,+2.027,-197.8,+21.302,50.35,0.00,0,49.87,0.00,0,50.06,0.00,0,51.15,0.00,0,50.06,0.00,0,51.15,0.00,0,53.41,0.00,0,97.12,0.00,1,50.95,0.00,0,,,-181.2,+8.330
38,OGBN\\Arxiv,LOP-GPN,57.74,0.00,0,63.37,0.00,0,0.00,0.00,1,67.00,0.00,0,66.77,0.00,0,67.00,0.00,0,68.12,0.00,0,66.28,0.00,0,67.44,0.00,0,54.58,0.00,0,+16.029,+17.598,-196.0,+7.011,57.49,0.00,0,56.06,0.00,0,67.29,0.00,0,82.46,0.00,0,67.29,0.00,0,73.03,0.00,0,68.39,0.00,0,92.39,0.00,0,87.08,0.00,0,+22.651,+17.534,-183.4,+49.155,57.85,0.00,0,57.93,0.00,0,65.65,0.00,1,70.58,0.00,1,65.65,0.00,1,68.03,0.00,1,63.98,0.00,1,72.86,0.00,0,64.08,0.00,1,,,-194.5,+14.350
39,OGBN\\Arxiv,CUQ-PPR,69.62,0.00,1,74.38,0.00,1,0.00,0.00,1,66.90,0.00,0,68.58,0.00,0,66.90,0.00,0,68.58,0.00,0,66.73,0.00,0,68.32,0.00,0,60.96,0.00,0,+43.309,+43.309,-192.8,+17.009,52.17,0.00,0,11.09,0.00,0,80.54,0.00,1,86.16,0.00,1,80.54,0.00,1,86.12,0.00,1,93.59,0.00,1,93.25,0.00,1,93.54,0.00,1,+67.299,+65.806,-172.8,+3285.7,69.21,0.00,1,67.42,0.00,1,54.99,0.00,0,56.42,0.00,0,54.99,0.00,0,56.41,0.00,0,59.85,0.00,0,60.33,0.00,0,59.56,0.00,0,+14.342,+14.315,-195.1,+110.38
40,OGBN\\Arxiv,CUQ-GCN,68.46,0.00,0,73.56,0.00,0,0.00,0.00,1,66.79,0.00,0,68.29,0.00,0,66.79,0.00,0,68.28,0.00,0,67.84,0.00,0,68.98,0.00,0,67.82,0.00,0,+43.554,+43.546,-191.1,+74.168,30.20,0.00,0,11.95,0.00,0,67.25,0.00,0,69.93,0.00,0,67.25,0.00,0,69.85,0.00,0,75.65,0.00,0,75.00,0.00,0,75.65,0.00,0,+32.099,+28.779,-180.8,+1042.4,67.74,0.00,0,65.95,0.00,0,53.50,0.00,0,54.23,0.00,0,53.50,0.00,0,54.23,0.00,0,55.39,0.00,0,55.47,0.00,0,55.39,0.00,0,+11.057,+11.047,-196.9,+57.279
41,OGBN\\Arxiv,CUQ-GAT,68.41,0.00,0,73.21,0.00,0,0.00,0.00,1,67.16,0.00,0,69.05,0.00,1,67.16,0.00,0,69.05,0.00,1,68.40,0.00,1,69.65,0.00,1,68.40,0.00,1,+44.057,+44.057,-191.3,+40.557,62.11,0.00,0,53.14,0.00,0,57.09,0.00,0,58.34,0.00,0,57.09,0.00,0,58.33,0.00,0,61.55,0.00,0,60.92,0.00,0,61.55,0.00,0,+16.077,+15.860,-194.7,+468.13,68.07,0.00,0,66.60,0.00,0,52.08,0.00,0,52.68,0.00,0,52.08,0.00,0,52.68,0.00,0,52.87,0.00,0,53.50,0.00,0,52.89,0.00,0,+5.844,+5.844,-198.4,+9.316
\end{filecontents*}
\csvreader[
        column count=99,
        tabular={c r | rrrrrr | rrrrrr},
        separator=comma,
        table head={%
            & \multicolumn{1}{c}{} &%
            \multicolumn{6}{c}{Leave-out Classes} &%
            \multicolumn{6}{c}{$\mathbf{x}^{(v)} \sim \mathcal{N}(0,1)$}%
            \\%
            & \multicolumn{1}{r}{} &%
            \multicolumn{1}{c}{\textbf{ID}} & %
            \multicolumn{5}{c}{\textbf{OOD-AUC-ROC}} &%
            \multicolumn{1}{c}{\textbf{ID}} & %
            \multicolumn{5}{c}{\textbf{OOD-AUC-ROC}} \\
            & \multicolumn{1}{r}{} &%
            \multicolumn{1}{c}{\textbf{Acc}} & %
            \multicolumn{1}{c}{$\mathrm{TU}$} & \multicolumn{1}{c}{$\mathrm{AU}$} & \multicolumn{1}{c}{$\mathrm{EU}$} & \multicolumn{1}{c}{$\mathrm{EU}_{\mathrm{PC}}$} & \multicolumn{1}{c}{$\mathrm{EU}_{\mathrm{SO}}$} &%
            \multicolumn{1}{c}{\textbf{Acc}} & %
            \multicolumn{1}{c}{$\mathrm{TU}$} & \multicolumn{1}{c}{$\mathrm{AU}$} & \multicolumn{1}{c}{$\mathrm{EU}$} & \multicolumn{1}{c}{$\mathrm{EU}_{\mathrm{PC}}$} & \multicolumn{1}{c}{$\mathrm{EU}_{\mathrm{SO}}$} %
            \\\toprule%
        },
        before reading={\setlength{\tabcolsep}{3pt}},
        table foot=\bottomrule,
        head to column names,
        late after line={\ifthenelse{\equal{\model}{APPNP}\and\not\equal{\id}{0}}{\\\midrule}{\\}},
        filter={\not\(\equal{\dataset}{OGBN\\Arxiv}\and\equal{\model}{Matern-GGP}\)}
    ]{tables/id_ood.csv}{}{%
        \ifthenelse{\equal{\model}{APPNP}}{%
            \ifthenelse{\equal{\dataset}{OGBN\\Arxiv}}{%
                \multirow{7}{*}[0em]{\shortstack{\dataset}}%
            }{\multirow{7}{*}[0em]{\shortstack{\dataset}}}%
        }{} &%
        \textbf{\model} &%
        \rawres{\oodLocIdAcc}{\oodLocIdAccSE}{\oodLocIdAccBest} &%
        \rawres{\oodLocOodTotalEntropy}{\oodLocOodTotalEntropySE}{\oodLocOodTotalEntropyBest} & %
        \rawres{\oodLocOodAleatoricEntropy}{\oodLocOodAleatoricEntropySE}{\oodLocOodAleatoricEntropyBest} & %
        \rawres{\oodLocOodEpistemicEntropyDiff}{\oodLocOodEpistemicEntropyDiffSE}{\oodLocOodEpistemicEntropyDiffBest} & %
        \rawres{\oodLocOodEpistemic}{\oodLocOodEpistemicSE}{\oodLocOodEpistemicBest} & %
        \rawres{\oodLocOodEpistemicEntropy}{\oodLocOodEpistemicEntropySE}{\oodLocOodEpistemicEntropyBest} %
        &%
        \rawres{\oodNormalIdAcc}{\oodNormalIdAccSE}{\oodNormalIdAccBest} &%
        \rawres{\oodNormalOodTotalEntropy}{\oodNormalOodTotalEntropySE}{\oodNormalOodTotalEntropyBest} & %
        \rawres{\oodNormalOodAleatoricEntropy}{\oodNormalOodAleatoricEntropySE}{\oodNormalOodAleatoricEntropyBest} & %
        \rawres{\oodNormalOodEpistemicEntropyDiff}{\oodNormalOodEpistemicEntropyDiffSE}{\oodNormalOodEpistemicEntropyDiffBest} & %
        \rawres{\oodNormalOodEpistemic}{\oodNormalOodEpistemicSE}{\oodNormalOodEpistemicBest} & %
        \rawres{\oodNormalOodEpistemicEntropy}{\oodNormalOodEpistemicEntropySE}{\oodNormalOodEpistemicEntropyBest} %
    }}
\end{table}
One practical application of \ac{uq} is the detection of outliers, i.e., distinguishing nodes that are in the data distribution from those that lie \acf{ood}.
Similar to \citet{stadler2021}, we generate outlier nodes in two ways:
\begin{enumerate}
    \item Nodes belonging to a pre-selected subset of classes are omitted from the training data; during testing, a model should then detect nodes from the unseen classes as outliers.
    \item The features of a random subset of test nodes are altered by adding Gaussian noise to them; a model should detect the noisy nodes as outliers.
\end{enumerate}
The performance of a given uncertainty measure is assessed via the \acf{auroc}.
\Cref{tbl:ood} shows the \ac{ood} detection performances and the \ac{id} accuracies on both \ac{ood} detection scenarios.
Note that standard errors were omitted because they are close to zero for all entries.

Looking at the \ac{id} accuracies, \ac{appnp} generally performs best in the leave-out classes scenario, while \ac{lopgpn} performs best in the Gaussian noise scenario.
Note, however, that the difference between the \ac{id} accuracies of \ac{appnp} and CUQ-PPR in the leave-out classes scenario is relatively small; this similarity is plausible since CUQ-PPR can be seen as a second-order, uncertainty-aware variant of \ac{appnp}.

Comparing the \ac{ood} detection performance of different models, \ac{cuqgnn}, more specifically CUQ-PPR, achieves the best \ac{auroc} values most often in the Gaussian noise scenario, followed by \ac{gpn} and \ac{lopgpn}.
In the leave-out classes scenario \ac{gpn} achieves the best result most often, followed by CUQ-GAT and \ac{lopgpn}.
This shows that, depending on the setting, both, the behavioral pooling of \ac{cuqgnn} and the axiomatic approaches of (\acs{lop}-)\acs{gpn} are well-suited for outlier detection problems.
To summarize, \textbf{the well-behaved \acp{arc} of \acp{cuqgnn} observed in the previous section also translate into a good practical performance on OOD tasks}.

\section{Conclusion}\label{sec:conclusion}

In this paper, we proposed the family of \ac{cuqgnn} models for uncertainty quantification in node classification.
Unlike \ac{gpn} and \ac{lopgpn}, which employ static axiomatic pooling schemes, \ac{cuqgnn} is based on the more flexible notion of behavioral pooling from the field of risk and decision analysis.
More specifically, behavioral pooling is realized by combining standard graph convolution operators with a \ac{postnet} model.
We demonstrate the effectiveness of our approach empirically by comparing it to multiple state-of-the-art baseline methods.

We envision three lines of future research.
First, a deeper and more systematic investigation of the intersection of opinion pooling and uncertainty propagation on graphs would be desirable.
While our proposed behavioral pooling approach achieves strong empirical results, it does not provide theoretical guarantees for its uncertainty estimates.
Second, as we have seen, choosing an appropriate graph convolution operator for \ac{cuqgnn} is a complex domain-dependent problem.
Therefore, it would be interesting to design an AutoML system to automatically configure \ac{cuqgnn} for a given domain.
Third, \ac{gpn}, \ac{lopgpn} and \ac{cuqgnn} are all only applicable to node classification tasks.
Extending this line of work to graph classification or node regression tasks would be interesting as well.

\begin{credits}

\subsubsection{\discintname}
The authors have no competing interests to declare that are
relevant to the content of this article.
\end{credits}
%
%
%

\begin{thebibliography}{49}
\providecommand{\natexlab}[1]{#1}
\providecommand{\url}[1]{\texttt{#1}}
\providecommand{\urlprefix}{URL }
\expandafter\ifx\csname urlstyle\endcsname\relax
  \providecommand{\doi}[1]{doi:\discretionary{}{}{}#1}\else
  \providecommand{\doi}{doi:\discretionary{}{}{}\begingroup \urlstyle{rm}\Url}\fi

\bibitem[{Abbas(2009)}]{abbas2009}
Abbas, A.E.: A {{Kullback-Leibler View}} of {{Linear}} and {{Log-Linear Pools}}. Decision Analysis \textbf{6}(1), 25--37 (2009)

\bibitem[{Albert and Barab{\'a}si(2002)}]{albert2002}
Albert, R., Barab{\'a}si, A.L.: Statistical mechanics of complex networks. Rev. Mod. Phys. \textbf{74}(1), 47--97 (2002)

\bibitem[{Arrow(1951)}]{arrow1951}
Arrow, K.J.: Social {{Choice}} and {{Individual Values}}. Wiley: New York (1951)

\bibitem[{Barab{\'a}si and Albert(1999)}]{barabasi1999}
Barab{\'a}si, A.L., Albert, R.: Emergence of {{Scaling}} in {{Random Networks}}. Science \textbf{286}(5439), 509--512 (1999)

\bibitem[{Bengs et~al.(2022)Bengs, H{\"u}llermeier, and Waegeman}]{bengs2022}
Bengs, V., H{\"u}llermeier, E., Waegeman, W.: Pitfalls of {{Epistemic Uncertainty Quantification}} through {{Loss Minimisation}}. {{NeurIPS}} \textbf{35}, 29205--29216 (2022)

\bibitem[{Bengs et~al.(2023)Bengs, H{\"u}llermeier, and Waegeman}]{bengs2023}
Bengs, V., H{\"u}llermeier, E., Waegeman, W.: On {{Second-Order Scoring Rules}} for {{Epistemic Uncertainty Quantification}}. In: {{ICML}}, pp. 2078--2091, PMLR (2023)

\bibitem[{Bilo{\v s} et~al.(2019)Bilo{\v s}, Charpentier, and G{\"u}nnemann}]{bilos2019}
Bilo{\v s}, M., Charpentier, B., G{\"u}nnemann, S.: Uncertainty on {{Asynchronous Time Event Prediction}}. In: {{NeurIPS}}, vol.~32 (2019)

\bibitem[{Bojchevski and G{\"u}nnemann(2018)}]{bojchevski2018}
Bojchevski, A., G{\"u}nnemann, S.: Deep {{Gaussian Embedding}} of {{Graphs}}: {{Unsupervised Inductive Learning}} via {{Ranking}}. In: {{ICLR}} (2018)

\bibitem[{Bronevich and Klir(2008)}]{bronevich2008}
Bronevich, A., Klir, G.J.: Axioms for uncertainty measures on belief functions and credal sets. In: {{NAFIPS}} 2008, pp. 1--6 (2008)

\bibitem[{Charpentier et~al.(2020)Charpentier, Z{\"u}gner, and G{\"u}nnemann}]{charpentier2020}
Charpentier, B., Z{\"u}gner, D., G{\"u}nnemann, S.: Posterior network: Uncertainty estimation without {{OOD}} samples via density-based pseudo-counts. In: {{NeurIPS}} (2020)

\bibitem[{Clemen and Winkler(1999)}]{clemen1999}
Clemen, R.T., Winkler, R.L.: Combining {{Probability Distributions From Experts}} in {{Risk Analysis}}. Risk Anal \textbf{19}(2), 187--203 (1999)

\bibitem[{Clemen and Winkler(2007)}]{clemen2007}
Clemen, R.T., Winkler, R.L.: Aggregating {{Probability Distributions}}. In: Advances in {{Decision Analysis}}: {{From Foundations}} to {{Applications}}, pp. 154--176 (2007)

\bibitem[{Damke and H{\"u}llermeier(2024)}]{damke2024}
Damke, C., H{\"u}llermeier, E.: Linear {{Opinion Pooling}} for {{Uncertainty Quantification}} on {{Graphs}}. In: UAI'2024, Barcelona (2024)

\bibitem[{Damke et~al.(2020)Damke, Melnikov, and H{\"u}llermeier}]{damke2020}
Damke, C., Melnikov, V., H{\"u}llermeier, E.: A {{Novel Higher-order Weisfeiler-Lehman Graph Convolution}}. In: {{ACML}}, pp. 49--64, PMLR (2020)

\bibitem[{Depeweg et~al.(2018)Depeweg, {Hernandez-Lobato}, {Doshi-Velez}, and Udluft}]{depeweg2018}
Depeweg, S., {Hernandez-Lobato}, J.M., {Doshi-Velez}, F., Udluft, S.: Decomposition of {{Uncertainty}} in {{Bayesian Deep Learning}} for {{Efficient}} and {{Risk-sensitive Learning}}. In: {{ICML}}, pp. 1184--1193, PMLR (2018)

\bibitem[{Duan et~al.(2024)Duan, Caffo, Bai, Sair, and Jones}]{duan2024}
Duan, R., Caffo, B., Bai, H.X., Sair, H.I., Jones, C.: Evidential {{Uncertainty Quantification}}: {{A Variance-Based Perspective}}. In: {{WACV}} (2024)

\bibitem[{Gasteiger et~al.(2018)Gasteiger, Bojchevski, and G{\"u}nnemann}]{gasteiger2018}
Gasteiger, J., Bojchevski, A., G{\"u}nnemann, S.: Predict then {{Propagate}}: {{Graph Neural Networks}} meet {{Personalized PageRank}}. In: {{ICLR}} (2018)

\bibitem[{Genest(1984)}]{genest1984}
Genest, C.: A {{Characterization Theorem}} for {{Externally Bayesian Groups}}. The Annals of Statistics \textbf{12}(3), 1100--1105 (1984)

\bibitem[{Genest and Zidek(1986)}]{genest1986}
Genest, C., Zidek, J.V.: Combining {{Probability Distributions}}: {{A Critique}} and an {{Annotated Bibliography}}. Statistical Science \textbf{1}(1), 114--135 (1986)

\bibitem[{Getoor(2005)}]{getoor2005}
Getoor, L.: Link-based {{Classification}}. In: Bandyopadhyay, S., Maulik, U., Holder, L.B., Cook, D.J. (eds.) Advanced {{Methods}} for {{Knowledge Discovery}} from {{Complex Data}}, pp. 189--207, Springer (2005)

\bibitem[{Giles et~al.(1998)Giles, Bollacker, and Lawrence}]{giles1998}
Giles, C.L., Bollacker, K.D., Lawrence, S.: {{CiteSeer}}: An automatic citation indexing system. In: Third {{ACM}} Conference on {{Digital}} Libraries, pp. 89--98 (1998)

\bibitem[{Hu et~al.(2020)Hu, Fey, Zitnik, Dong, Ren, Liu, Catasta, and Leskovec}]{hu2020}
Hu, W., Fey, M., Zitnik, M., Dong, Y., Ren, H., Liu, B., Catasta, M., Leskovec, J.: Open {{Graph Benchmark}}: {{Datasets}} for {{Machine Learning}} on {{Graphs}}. In: {{NeurIPS}}, vol.~33, pp. 22118--22133 (2020)

\bibitem[{H{\"u}llermeier and Waegeman(2021)}]{hullermeier2021}
H{\"u}llermeier, E., Waegeman, W.: Aleatoric and epistemic uncertainty in machine learning: An introduction to concepts and methods. Mach Learn \textbf{110}(3) (2021)

\bibitem[{Huseljic et~al.(2021)Huseljic, Sick, Herde, and Kottke}]{huseljic2021}
Huseljic, D., Sick, B., Herde, M., Kottke, D.: Separation of {{Aleatoric}} and {{Epistemic Uncertainty}} in {{Deterministic Deep Neural Networks}}. In: {{ICPR}}'2020 (2021)

\bibitem[{Kendall and Gal(2017)}]{kendall2017}
Kendall, A., Gal, Y.: What uncertainties do we need in {{Bayesian}} deep learning for computer vision? In: {{NeurIPS}}, pp. 5580--5590, Curran Associates Inc. (2017)

\bibitem[{Kipf and Welling(2017)}]{kipf2017}
Kipf, T.N., Welling, M.: Semi-{{Supervised Classification}} with {{Graph Convolutional Networks}}. In: {{ICLR}} (2017)

\bibitem[{Koliander et~al.(2022)Koliander, {El-Laham}, Djuri{\'c}, and Hlawatsch}]{koliander2022}
Koliander, G., {El-Laham}, Y., Djuri{\'c}, P.M., Hlawatsch, F.: Fusion of {{Probability Density Functions}}. IEEE \textbf{110}(4), 404--453 (2022)

\bibitem[{Kopetzki et~al.(2021)Kopetzki, Charpentier, Z{\"u}gner, Giri, and G{\"u}nnemann}]{kopetzki2021}
Kopetzki, A.K., Charpentier, B., Z{\"u}gner, D., Giri, S., G{\"u}nnemann, S.: Evaluating {{Robustness}} of {{Predictive Uncertainty Estimation}}: {{Are Dirichlet-based Models Reliable}}? In: {{ICML}}, pp. 5707--5718, PMLR (2021)

\bibitem[{Kotelevskii et~al.(2023)Kotelevskii, Horv{\'a}th, Nandakumar, Tak{\'a}{\v c}, and Panov}]{kotelevskii2023}
Kotelevskii, N., Horv{\'a}th, S., Nandakumar, K., Tak{\'a}{\v c}, M., Panov, M.: Dirichlet-based {{Uncertainty Quantification}} for {{Personalized Federated Learning}} with {{Improved Posterior Networks}} (2023)

\bibitem[{Malinin and Gales(2018)}]{malinin2018}
Malinin, A., Gales, M.: Predictive uncertainty estimation via prior networks. In: {{NeurIPS}}, pp. 7047--7058, Curran Associates Inc. (2018)

\bibitem[{Maron et~al.(2019)Maron, {Ben-Hamu}, Serviansky, and Lipman}]{maron2019}
Maron, H., {Ben-Hamu}, H., Serviansky, H., Lipman, Y.: Provably {{Powerful Graph Networks}}. arXiv  (2019)

\bibitem[{McAuley et~al.(2015)McAuley, Targett, Shi, and {van den Hengel}}]{mcauley2015}
McAuley, J., Targett, C., Shi, Q., {van den Hengel}, A.: Image-{{Based Recommendations}} on {{Styles}} and {{Substitutes}}. In: 38th {{International ACM SIGIR Conference}} on {{Research}} and {{Development}} in {{Information Retrieval}}, pp. 43--52 (2015)

\bibitem[{McCallum et~al.(2000)McCallum, Nigam, Rennie, and Seymore}]{mccallum2000}
McCallum, A.K., Nigam, K., Rennie, J., Seymore, K.: Automating the {{Construction}} of {{Internet Portals}} with {{Machine Learning}}. Information Retrieval \textbf{3}(2) (2000)

\bibitem[{Namata et~al.(2012)Namata, London, Getoor, and Huang}]{namata2012}
Namata, G., London, B., Getoor, L., Huang, B.: Query-driven {{Active Surveying}} for {{Collective Classification}}. In: {{MLG Workshop}} (2012)

\bibitem[{Pal et~al.(1993)Pal, Bezdek, and Hemasinha}]{pal1993}
Pal, N.R., Bezdek, J.C., Hemasinha, R.: Uncertainty measures for evidential reasoning {{II}}: {{A}} new measure of total uncertainty. IJAR \textbf{8}(1), 1--16 (1993)

\bibitem[{Redner(1998)}]{redner1998}
Redner, S.: How popular is your paper? {{An}} empirical study of the citation distribution. Eur. Phys. J. B \textbf{4}(2), 131--134 (1998)

\bibitem[{Sale et~al.(2023)Sale, Hofman, Wimmer, H{\"u}llermeier, and Nagler}]{sale2023a}
Sale, Y., Hofman, P., Wimmer, L., H{\"u}llermeier, E., Nagler, T.: Second-{{Order Uncertainty Quantification}}: {{Variance-Based Measures}} (2023)

\bibitem[{Sen et~al.(2008)Sen, Namata, Bilgic, Getoor, Galligher, and {Eliassi-Rad}}]{sen2008}
Sen, P., Namata, G., Bilgic, M., Getoor, L., Galligher, B., {Eliassi-Rad}, T.: Collective {{Classification}} in {{Network Data}}. AI Magazine \textbf{29}(3), 93--93 (2008)

\bibitem[{Sensoy et~al.(2018)Sensoy, Kaplan, and Kandemir}]{sensoy2018}
Sensoy, M., Kaplan, L., Kandemir, M.: Evidential deep learning to quantify classification uncertainty. In: {{NeurIPS}}, pp. 3183--3193, Curran Associates Inc. (2018)

\bibitem[{Shu et~al.(2017)Shu, Sliva, Wang, Tang, and Liu}]{shu2017}
Shu, K., Sliva, A., Wang, S., Tang, J., Liu, H.: Fake {{News Detection}} on {{Social Media}}: {{A Data Mining Perspective}}. SIGKDD Explor. Newsl. \textbf{19}(1), 22--36 (2017)

\bibitem[{Stadler et~al.(2021)Stadler, Charpentier, Geisler, Z{\"u}gner, and G{\"u}nnemann}]{stadler2021}
Stadler, M., Charpentier, B., Geisler, S., Z{\"u}gner, D., G{\"u}nnemann, S.: Graph {{Posterior Network}}: {{Bayesian Predictive Uncertainty}} for {{Node Classification}} (2021)

\bibitem[{Veli{\v c}kovi{\'c} et~al.(2018)Veli{\v c}kovi{\'c}, Cucurull, Casanova, Romero, Li{\`o}, and Bengio}]{velickovic2018}
Veli{\v c}kovi{\'c}, P., Cucurull, G., Casanova, A., Romero, A., Li{\`o}, P., Bengio, Y.: Graph {{Attention Networks}}. In: {{ICLR}} (2018)

\bibitem[{Wang et~al.(2008)Wang, Yu, and Yu}]{wang2008}
Wang, M., Yu, G., Yu, D.: Measuring the preferential attachment mechanism in citation networks. Physica A: Stat Mechanics and its Applications \textbf{387}(18) (2008)

\bibitem[{Wang et~al.(2019)Wang, Ji, Shi, Wang, Ye, Cui, and Yu}]{wang2019}
Wang, X., Ji, H., Shi, C., Wang, B., Ye, Y., Cui, P., Yu, P.S.: Heterogeneous {{Graph Attention Network}}. In: The {{World Wide Web Conference}}, pp. 2022--2032 (2019)

\bibitem[{Wimmer et~al.(2023)Wimmer, Sale, Hofman, Bischl, and H{\"u}llermeier}]{wimmer2023}
Wimmer, L., Sale, Y., Hofman, P., Bischl, B., H{\"u}llermeier, E.: Quantifying aleatoric and epistemic uncertainty in machine learning: {{Are}} conditional entropy and mutual information appropriate measures? In: {{UAI'2023}}, pp. 2282--2292, PMLR (2023)

\bibitem[{Xu et~al.(2018)Xu, Hu, Leskovec, and Jegelka}]{xu2018}
Xu, K., Hu, W., Leskovec, J., Jegelka, S.: How {{Powerful}} are {{Graph Neural Networks}}? In: {{ICLR}} (2018)

\bibitem[{Zhang et~al.(2019)Zhang, Song, Huang, Swami, and Chawla}]{zhang2019a}
Zhang, C., Song, D., Huang, C., Swami, A., Chawla, N.V.: Heterogeneous {{Graph Neural Network}}. In: 25th {{ACM SIGKDD International Conference}} on {{Knowledge Discovery}} \& {{Data Mining}}, pp. 793--803 (2019)

\bibitem[{Zhao et~al.(2020)Zhao, Chen, Hu, and Cho}]{zhao2020}
Zhao, X., Chen, F., Hu, S., Cho, J.H.: Uncertainty aware semi-supervised learning on graph data. In: {{NeurIPS}}, pp. 12827--12836, Curran Associates Inc. (2020)

\bibitem[{Zhu et~al.(2020)Zhu, Yan, Zhao, Heimann, Akoglu, and Koutra}]{zhu2020}
Zhu, J., Yan, Y., Zhao, L., Heimann, M., Akoglu, L., Koutra, D.: Beyond homophily in graph neural networks: Current limitations and effective designs. In: {{NeurIPS}}, pp. 7793--7804, Curran Associates Inc. (2020)

\end{thebibliography}

\end{document}